\documentclass[runningheads]{llncs}
\usepackage{graphicx}
\usepackage[table]{xcolor}

\usepackage{amsmath,amsfonts,bm}

\newcommand{\cB}{\mathbf{c}}

\def\eqref#1{equation~\ref{#1}}

\def\1{\bm{1}}

\newcommand{\test}{\mathcal{D_{\mathrm{test}}}}

\DeclareMathAlphabet{\mathsfit}{\encodingdefault}{\sfdefault}{m}{sl}
\SetMathAlphabet{\mathsfit}{bold}{\encodingdefault}{\sfdefault}{bx}{n}

\newcommand{\E}{\mathbb{E}}

\newcommand{\R}{\mathbb{R}}

\newcommand{\matr}[1]{\mathbf{#1}} 

\usepackage{float} 
\usepackage{wrapfig}

\usepackage{tikz}
\usepackage{comment}
\usepackage{amsmath,amssymb} 
\usepackage{color}
\usepackage{algorithm,algpseudocode}
\usepackage{booktabs} 
\usepackage{multirow}
\newlength{\Oldarrayrulewidth}

\newcommand{\etal}{\textit{et al}. }
\newcommand{\ie}{\textit{i}.\textit{e}.}
\newcommand{\eg}{\textit{e}.\textit{g}.} 

\usepackage[accsupp]{axessibility}  

\usepackage[bookmarksnumbered=true]{hyperref} 

\hypersetup{
     colorlinks = true,
     linkcolor = black,
     anchorcolor = black,
     citecolor = blue,
     filecolor = black,
     urlcolor = purple
     }
\usepackage{breakcites}
 
\begin{document}
\pagestyle{headings}
\mainmatter
\def\ECCVSubNumber{599}  

\title{Neural Architecture Search for \\ Spiking Neural Networks} 

\titlerunning{Neural Architecture Search for SNNs}
\authorrunning{Y. Kim, Y. Li, H. Park, V. Yeshwanth, P. Panda.}

\author{Youngeun Kim, Yuhang Li, Hyoungseob Park, \\ Yeshwanth Venkatesha, and Priyadarshini Panda}
\institute{
Department of Electrical Engineering\\
Yale University\\
New Haven, CT, USA\\
\email{\scriptsize \{youngeun.kim, yuhang.li, hyoungseob.park,  yeshwanth.venkatesha, priya.panda\}@yale.edu}
}

\maketitle

\begin{abstract}
Spiking Neural Networks (SNNs) have gained huge attention as a potential energy-efficient alternative to conventional Artificial Neural Networks (ANNs) due to their inherent high-sparsity activation.
However, most prior SNN methods use ANN-like architectures (\eg, VGG-Net or ResNet), which could provide sub-optimal performance for temporal sequence processing of binary information in SNNs.
To address this, in this paper, we introduce a novel Neural Architecture Search~(NAS) approach for finding better SNN architectures.
Inspired by recent NAS approaches that find the optimal architecture from activation patterns at initialization, we select the architecture that can represent diverse spike activation patterns across different data samples without training. 
Moreover, to further leverage the temporal information among the spikes, we search for feed-forward connections as well as backward connections (\ie, temporal feedback connections) between layers.
Interestingly, SNASNet found by our search algorithm achieves higher performance with backward connections, demonstrating the importance of designing SNN architecture for suitably using temporal information.
We conduct extensive experiments on three image recognition benchmarks where we show that SNASNet achieves state-of-the-art performance with significantly lower timesteps (5 timesteps). 
Code is available on \href{https://github.com/Intelligent-Computing-Lab-Yale/Neural-Architecture-Search-for-Spiking-Neural-Networks}{Github}. 
\end{abstract}

\section{Introduction}

Spiking Neural Networks (SNNs) \cite{roy2019towards,christensen20222022,wu2018spatio,wu2019direct,kundu2021hire,fang2021deep} have gained increasing attention as a promising paradigm for low-power intelligence.
Inspired by biological neuronal functionality, SNNs process visual information with binary spikes over multiple timesteps.
So far, the majority of works on SNNs have focused on image classification problem \cite{roy2019towards} to develop an energy-efficient alternative to Artificial Neural Networks (ANNs). 
To this end, recent SNN works utilize ANN architectures (\eg, VGG-Net \cite{simonyan2014very} or ResNet \cite{he2016deep}) designed by human experts. 
While SNNs show an impressive advantage on energy-efficiency, they still lag behind ANNs in terms of accuracy.

\begin{figure}[t]
\begin{center}
\def\arraystretch{0.5}
\begin{tabular}{@{}c@{}c@{}c@{}c@{}c}

\includegraphics[width=0.34\linewidth]{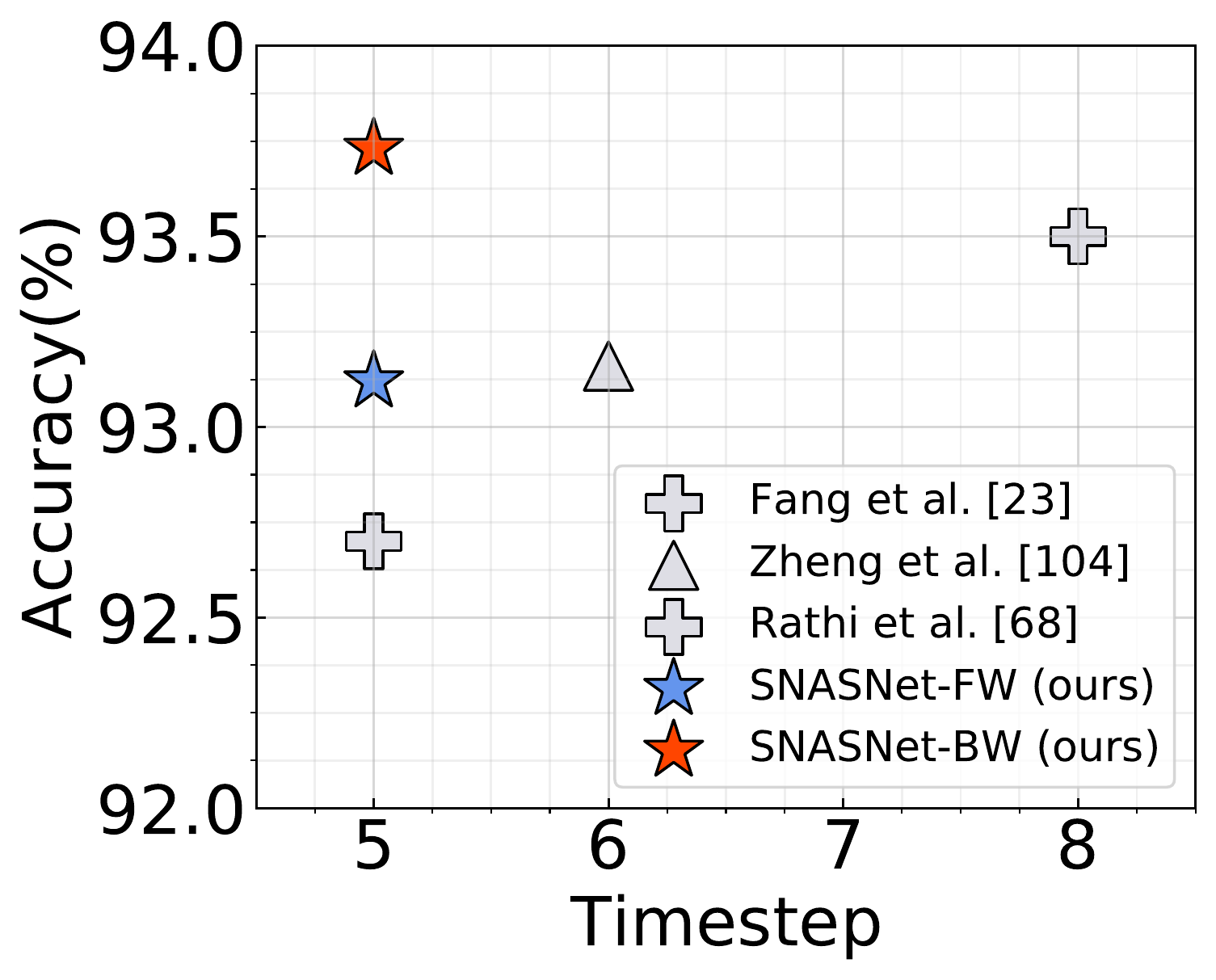} &
\includegraphics[width=0.325\linewidth]{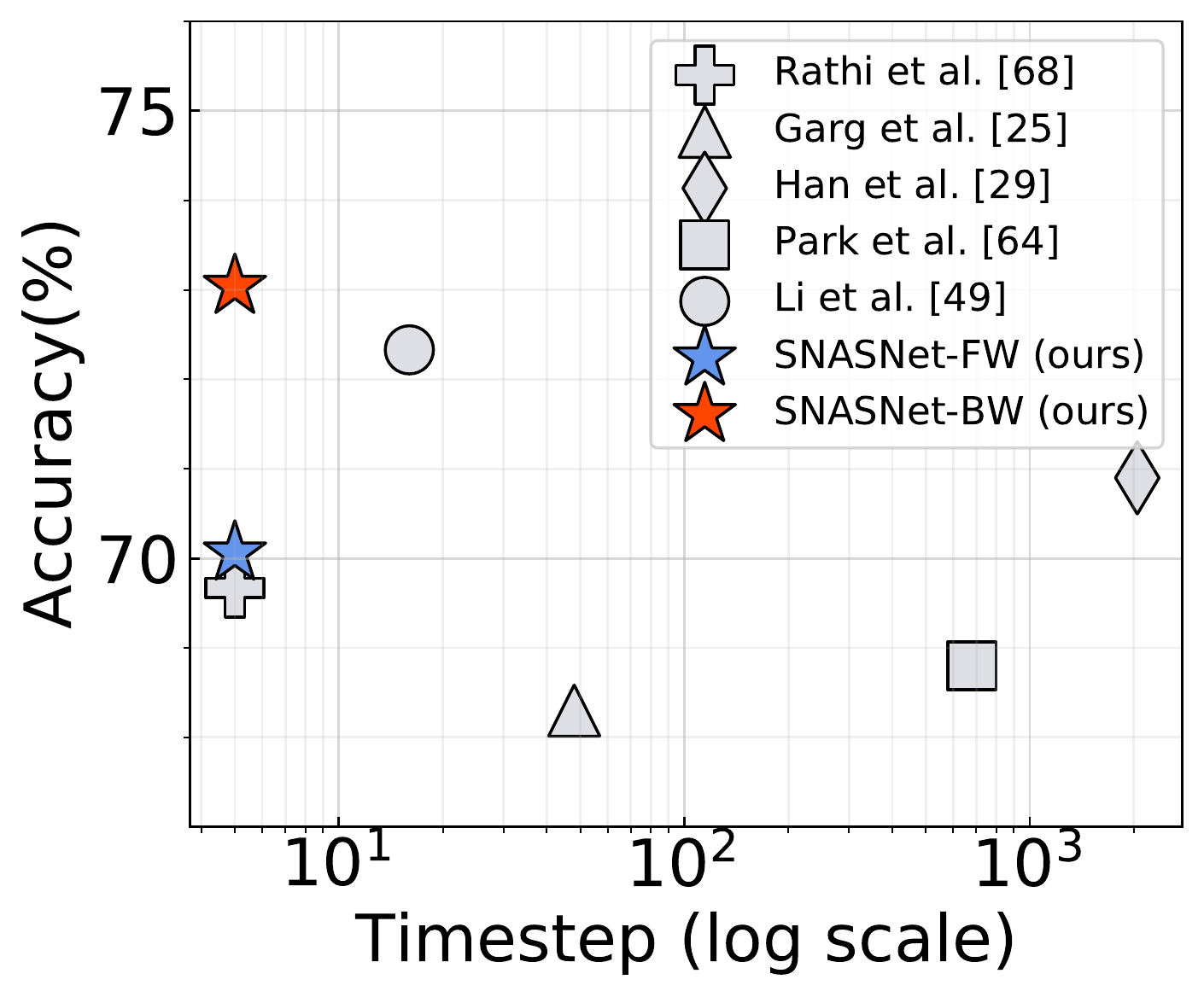} &
\includegraphics[width=0.325\linewidth]{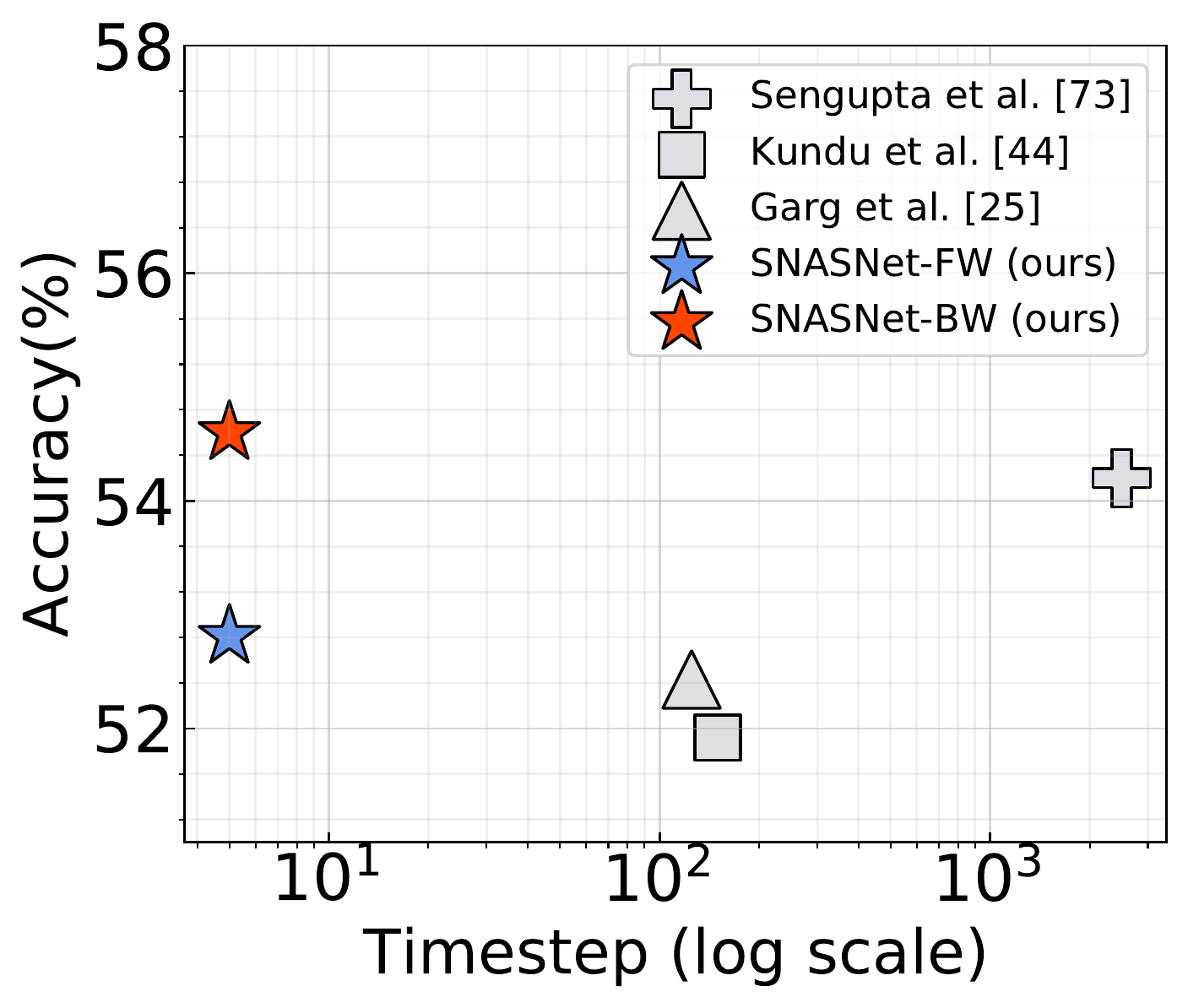} 
\\
 {(a) CIFAR10 } & 
{(b) CIFAR100 } & {(c) TinyImageNet} \\
\end{tabular}
\caption{ 
Accuracy and timesteps for different SNN models on (a) CIFAR10, (b) CIFAR100, and (c) TinyImageNet dataset. While showing comparable accuracy with state-of-the-art networks, SNASNet achieves significantly lower latency. Also, SNASNet-Bw where we search both forward and backward connections provides better performance than the SNASNet-Fw with only forward connections.}
\label{fig:intro:performance}
\end{center}
\end{figure}

In this paper, we argue that the inherent structural/functional difference between ANNs and SNNs induces an unignorable architectural gap, resulting in a sub-optimal solution when we naively deploy ANN architectures on SNNs.
Specifically, different from ANNs with ReLU neurons, SNNs consist of Leaky Integrate-and-Fire (LIF) neurons which store and transmit temporal information.
However, manually searching for SNN-friendly architectures is
laborious.
Therefore, we use Neural Architecture Search (NAS) \cite{zoph2016neural,zoph2018learning,cai2018proxylessnas,hu2020dsnas,shu2019understanding,zhang2021neural}, which can automatically discover the optimal SNN architecture.
Although NAS has become a prevalent technique in various ANN tasks \cite{chen2019detnas,chen2021glit,gong2019autogan,zhang2021dcnas}, NAS for designing SNNs has not been investigated. 
In this work, we ask two questions: \\
\textit{{Q1.    Which NAS algorithm is suitable for SNNs?}}\\
\textit{{Q2.  Which SNN architecture provides  better performance on an image recognition task?}
}

For the first question, we highlight that the mainstream NAS algorithms either require multiple training stages \cite{zoph2016neural,zoph2018learning,baker2016designing,zhong2018practical,tan2019mnasnet} or require training a supernet once with all architecture candidates \cite{liu2018darts,guo2020single,xie2018snas,cai2018proxylessnas} which takes longer training time to converge than standard training.
As SNNs have a significantly slower training process compared to ANNs (\eg, training SNN with timestep 10 on MNIST with NVIDIA V100 GPU takes 11.43$\times$ more latency compared to the same ANN architecture \cite{liang2021h2learn}), the above NAS approaches are difficult to be applied on SNNs.
On the other hand, recent works \cite{mellor2021neural,xu2021knas,chen2021neural} have proposed efficient NAS approaches that search the best neuron cell from initialized networks without any training. 
Specifically, \cite{mellor2021neural} shows that the network architecture with a high representation power at initialization is likely to achieve higher post-training accuracy.
Motivated by this, without the training process, we select the SNN architecture that can represent diverse spike activation patterns across different data samples. To quantify the diversity of networks, we measure the distance of temporal activation patterns between different mini-batch samples. However, SNNs show high sparsity variation on the temporal patterns across different mini-batches, resulting in inaccurate distance measures.
To address this, we normalize the distance measure based on the sparsity of given activation patterns, that we term as Sparsity-Aware Hamming Distance (SAHD).

To answer the second question, we search the optimal architecture block for SNNs.
Here, we find the connection topology as well as the corresponding operation for each connection following previous works \cite{dong2020bench,ying2019bench}. 
Different from ANNs, SNNs can leverage backward connections as they convey information through time.
The backward connections in SNNs can compute more efficiently because each neuron can participate several times in a network computation \cite{bellec2020solution}, 
and they are likely to capture the temporal correlation of the given input.
A line of work has studied backward connections in SNNs with various architectures and training methods \cite{demin2018recurrent,zhang2019spike,bellec2020solution,jia2021neuronal,panda2017learning,panda2020toward}.
Therefore, we search backward connections as well as forward connections through our NAS algorithm.
Surprisingly, SNNs with backward connections yield improved accuracy by up to 3$\%$ across various benchmark datasets compared to SNNs with forward connections only.
Also, as shown in Fig. \ref{fig:intro:performance},  \textit{SNASNet} founded by our NAS algorithm achieves state-of-the-art performance with a significantly small number of timesteps.

In summary, our key contributions are as follows: 
(1) So far, most SNN literature deploys architectures from ANN models which can yield sub-optimal performance for SNNs.
For the first time, we showcase a NAS technique for finding better SNN architecture on the image recognition task.
(2) Motivated by the prior work \cite{mellor2021neural,xu2021knas,chen2021neural}, we find an SNN-friendly architecture by comparing temporal activation without any training process.
Eliminating the training cost to find the optimal architecture brings a huge advantage for SNNs that require significantly longer training time compared to ANNs.
(3) We also propose Sparsity-Aware Hamming Distance (SAHD) for addressing sparsity variation of LIF neurons. 
(4) Furthermore, we search backward connections for leveraging temporal information in spiking inputs, which has not been explored before in NAS approaches for ANN architecture.

\section{Related Work}

\subsection{Spiking Neural Networks}

Spiking Neural Networks (SNNs) have gained great attention as an energy-efficient alternative over standard Artificial Neural Networks (ANNs) \cite{roy2019towards,panda2020toward,cao2015spiking,diehl2015unsupervised,comsa2020temporal,mostafa2017supervised,sengupta2019going,han2020rmp,diehl2015fast,rueckauer2017conversion,yousefzadeh2018practical,jin2010implementing,kundu2021hire,yao2021temporal,li2021free,li2022converting,kim2020revisiting,kim2021visual,kim2021optimizing,kundu2021hire,kim2021privatesnn,venkatesha2021federated,li2021differentiable,deng2022temporal}. 
SNNs process temporal information through weight connections and a Leak-Integrate-and-Fire (LIF) neuron  \cite{izhikevich2003simple} which works as a non-linear activation in SNNs.
The LIF neuron has its own memory called membrane potential that can store the temporal spike dynamics by accumulating incoming spike signals.
If the membrane potential exceeds a firing threshold, the neuron generates a post-synaptic spike.
The integrate-and-fire behavior of neurons induces non-differentiable transfer function. As a result, standard backpropagation is difficult to be applied during the training phase \cite{neftci2019surrogate}.

To address this, various methods have been proposed to circumvent the non-differentiable backpropagation problem. Among them, surrogate gradient learning approaches
have become popular \cite{lee2016training,lee2020enabling,neftci2019surrogate,shrestha2018slayer,gu2019stca,wu2018spatio} due to their higher performance and smaller number of timesteps compared to other training techniques.
They define a surrogate function for LIF neurons when calculating backward gradients.
Wu \etal \cite{wu2019direct} represent the LIF model in a discrete-time domain and enable SNN training with a Pytorch platform.
The authors of \cite{wu2021training} propose
a training algorithm that calculates backward gradients of the accumulated input and output spikes over the time window.
Tandem learning \cite{wu2019tandem,wu2020progressive}
utilizes an auxiliary ANN that facilitates stable error back-propagation for SNN training.
A line of work \cite{fang2021incorporating,rathi2021diet} train membrane decay or firing threshold in an LIF neuron, which improves the representation power of SNNs. 
Also, Batch Norm (BN)  \cite{ioffe2015batch}  has been applied to accelerate the training process of SNNs \cite{ledinauskas2020training,kim2020revisiting,zheng2020going}. 
In spite of the recent developments in SNN training techniques, all of the prior methods leverage ANN architecture, such as, VGG and ResNet families.
We assert that these architectures may provide sub-optimal solution for SNNs.
Different from previous methods, we search better SNN architectures for the image recognition task which has not been explored so far.
We notify that the concurrent work \cite{na2022autosnn} also aims to find SNN-friendly architecture using evolutionary algorithm, whereas our work is based on NAS without training technique. 
We also have different search spaces from \cite{na2022autosnn} where we more focus on searching backward connections.

\subsection{Neural Architecture Search}

Neural Architecture Search (NAS) has been proposed to discover high-performing networks  \cite{zoph2016neural,zoph2018learning,cai2018proxylessnas,hu2020dsnas,shu2019understanding,zhang2021neural}.
The early stage of NAS algorithm uses reinforcement learning \cite{zoph2016neural,zoph2018learning,baker2016designing,zhong2018practical,tan2019mnasnet} or evolutionary algorithm  \cite{real2019regularized}.
However, such methods require training the searched architecture from scratch for each search step, which is extremely computationally expensive.
To address this, weight-sharing approaches have been proposed \cite{cai2018proxylessnas,liu2018darts,wu2019fbnet,xie2018snas,bender2018understanding,brock2017smash,pham2018efficient,zhang2020you,guo2020single,chen2021bn,yang2021towards}.
They train the supernet once which includes all architecture candidates.
For instance, Darts \cite{liu2018darts} jointly optimizes the network parameters and the importance of each architecture candidate.
Also, SPOS \cite{guo2020single} trains the weight parameters with uniform forward path sampling and finds the optimal architecture via evolutionary strategy.
The weight-sharing methods do not require training the architecture from scratch at each search step, resulting in better efficiency compared to previous NAS algorithms.
In very recent works, the key focus has been the efficiency of the NAS technique \cite{yang2021hournas,yang2021netadaptv2,yan2021fp,abdelfattah2021zero,zhao2021few} owing to the growing size of dataset and architecture.
Interestingly, a line of work suggests the concept of NAS without training where the networks do not require training during the search stage \cite{xu2021knas,mellor2021neural,chen2021neural}.
This can significantly reduce the computational cost for searching optimal architecture.
At the same time, several benchmarks \cite{dong2020bench,ying2019bench,siems2020bench,duan2021transnas} have been proposed in order to remove the burden of training time.
Following the success of NAS on image classification domain, NAS has been deployed on various tasks such as 
object detection \cite{chen2019detnas}, segmentation  \cite{liu2019auto,zhang2021dcnas}, GAN \cite{gong2019autogan}, transformer \cite{chen2021glit}, and human pose estimation \cite{zeng2021neural,xu2021vipnas}.
Despite the huge progress of NAS algorithm in ANN domain, NAS for SNNs has not been developed yet.
In this work, we aim to build better SNN architecture
by leveraging NAS.
Different from the previous methods that search only forward connections of the networks, we search for backward connections in addition to forward, which furthers leverage the temporal information of spikes.

\section{Preliminaries}

\subsection{Leaky Integrate-and-Fire neuron}

Leaky Integrate-and-Fire (LIF) neuron is widely used for constructing SNNs \cite{roy2019towards,wu2019direct,fang2021incorporating}.
A neuron has a membrane potential
that stores the temporal spike information.
We convert the above continuous differential equation into a discrete version as in previous works \cite{wu2019direct,fang2021incorporating}:
\begin{equation}
    u_i^t = (1 - \frac{1}{\tau_m})  u_i^{t-1} + \frac{1}{\tau_m} \sum_j w_{ij}o^t_j, 
    \label{eq:LIF}
\end{equation}
where, $u_{i}^{t}$  represents the membrane potential of a neuron $i$ at timestep $t$, $\tau_m$ is a time constant for decaying the membrane potential.
Also, $w_{ij}$ stands for weight connections between neuron $j$ and neuron $i$.
The neuron $i$ accumulates membrane potential and generates a spike output $o_i^{t}$ whenever membrane potential exceeds the threshold. After firing, the membrane potential is reset to zero.

\subsection{NAS without Training}

Compared to standard ANNs, SNNs require significantly higher  computational cost for training due to multiple feedforward steps  \cite{liang2021h2learn}.
This makes it difficult to search for an optimal SNN architecture with  NAS techniques that train the architecture candidate multiple times \cite{zoph2016neural,zoph2018learning,baker2016designing,zhong2018practical,tan2019mnasnet} or train a complex supernet  \cite{liu2018darts,guo2020single,xie2018snas,cai2018proxylessnas}.
To minimize the training budget, our work is motivated by the previous works \cite{mellor2021neural,xu2021knas,chen2021neural} which demonstrate that the optimal architecture can be founded without any training process.
Specifically, Mellor \etal \cite{mellor2021neural} provide the interesting observation that the architecture having distinctive representations across different data samples is likely to achieve higher post-training performance.
To measure the discriminative power of initialized networks, they utilize the activation pattern of ReLU neurons as a binary indicator.
If the ReLU neuron generates a positive value (\ie, input  $>0$), the neuron is mapped to $1$; otherwise $0$.
As a result, ReLU neurons in one layer can be encoded to binary vector $\cB$.
Given $N$ samples in a mini-batch, they construct a kernel matrix by computing Hamming distance $d_H(\cB_i, \cB_j)$ between different samples $i$ and $j$, which can be formulated as follows:
\begin{equation}
\matr{K}_H = 
\begin{pmatrix}
    N_A {-} d_H(\cB_1, \cB_1) & \cdots & N_A {-} d_H(\cB_1, \cB_N) \\
    \vdots & \ddots & \vdots \\
    N_A {-} d_H(\cB_N, \cB_1) & \cdots & N_A {-} d_H(\cB_N, \cB_N) \\
\end{pmatrix}
\label{eq:KH-wot}
\end{equation}
Here, $N_A$ stands for the number of ReLU neurons in the given layer.
The final score of the architecture candidate is obtained by:
\begin{equation}
    s =  \log(\det|\sum_{l}\matr{K}_H^l|),
    \label{eq:score-wot}
\end{equation}
where, $\matr{K}^l_H$ is the kernel matrix at layer $l$. 
A high score implies low off-diagonal elements of kernel matrix $\matr{K}_H$, which means that the activation patterns from different samples are not similar.
Finally, the highest-scored architecture among the candidates is selected for training.

\section{Methodology}

In this section, we first introduce a temporal binary indicator of an LIF neuron based on the concept of linear region in neural networks.
After that, we present sparsity-aware hamming distance that accounts for the sparsity variation of an LIF neuron.
Finally, we provide the search space for our NAS algorithm where we find both forward and backward connections.

\subsection{Linear Regions from LIF neurons}

NAS without training approaches in ANN domain \cite{mellor2021neural,chen2021neural} are based on the  theoretical concept of linear region in neural networks \cite{hanin2019deep,xiong2020number,hanin2019complexity,raghu2017expressive,montufar2014number}.
That is, each piecewise linear function (such as, ReLU) divides the input space into multiple linear regions.
The composition of multiple piecewise linear functions brings multiple linear regions on the input space.
Such a pattern of linear regions is used for measuring the representation power of initialized networks by comparing the patterns between different samples.
Here, based on previous work, we introduce the definition of neuron transition (\ie, the boundary of linear region) in a piecewise linear function.

\noindent
\textbf{Definition 1.} (Raghu \etal \cite{raghu2017expressive}) 
\textit{For fixed W, we say a neuron with piecewise
linear region transitions between inputs x, $x + \delta$ if its activation function switches linear region between $x$ and $x + \delta$.}

\begin{figure}[t]
\noindent\begin{minipage}{\textwidth}
\begin{minipage}{0.49\textwidth}
\centering%
\begin{figure}[H]
\begin{center}
\def\arraystretch{0.5}
\begin{tabular}{@{}c@{}c@{}c@{}c@{}}
\includegraphics[width=0.45\linewidth]{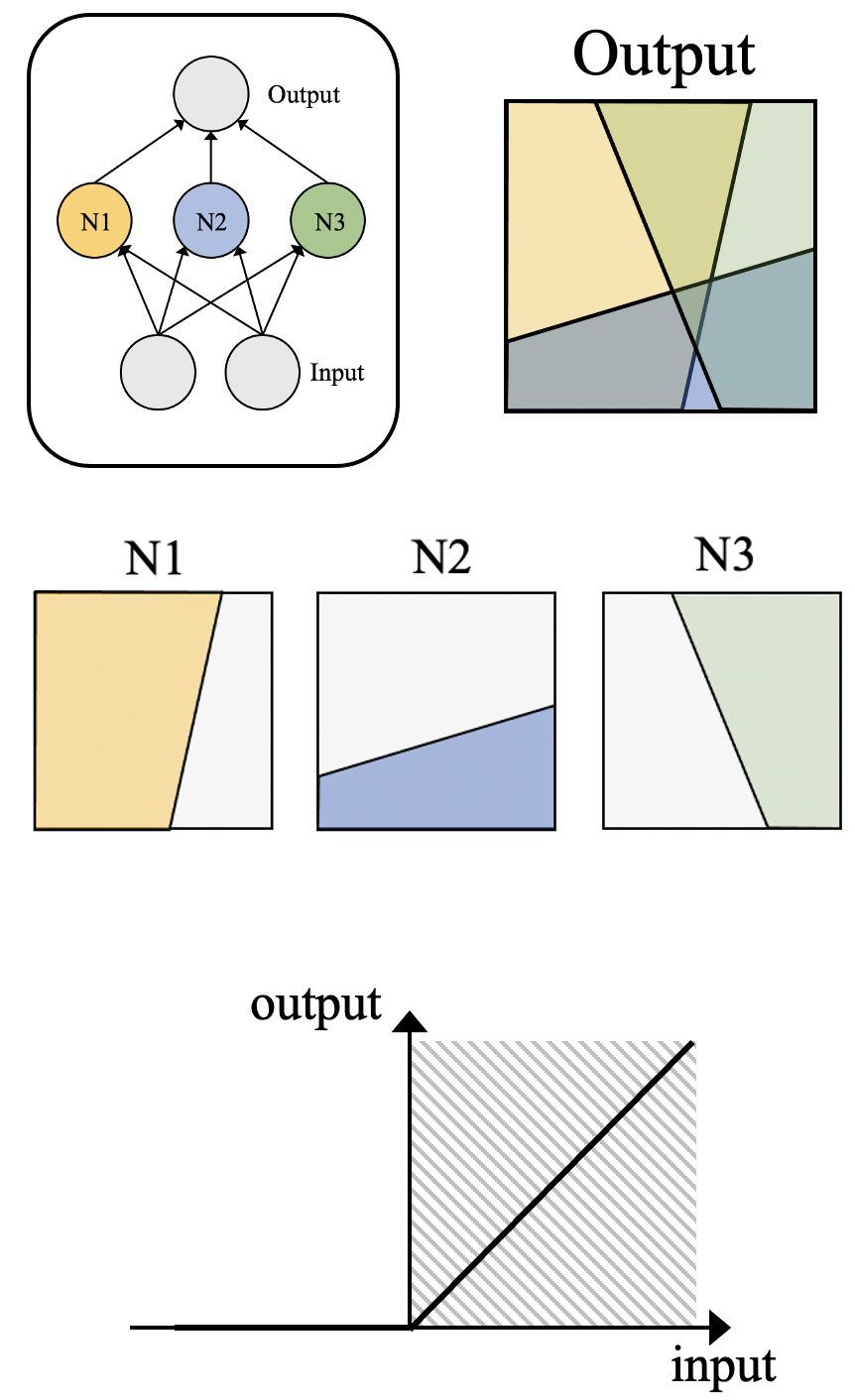} &
\includegraphics[width=0.56\linewidth]{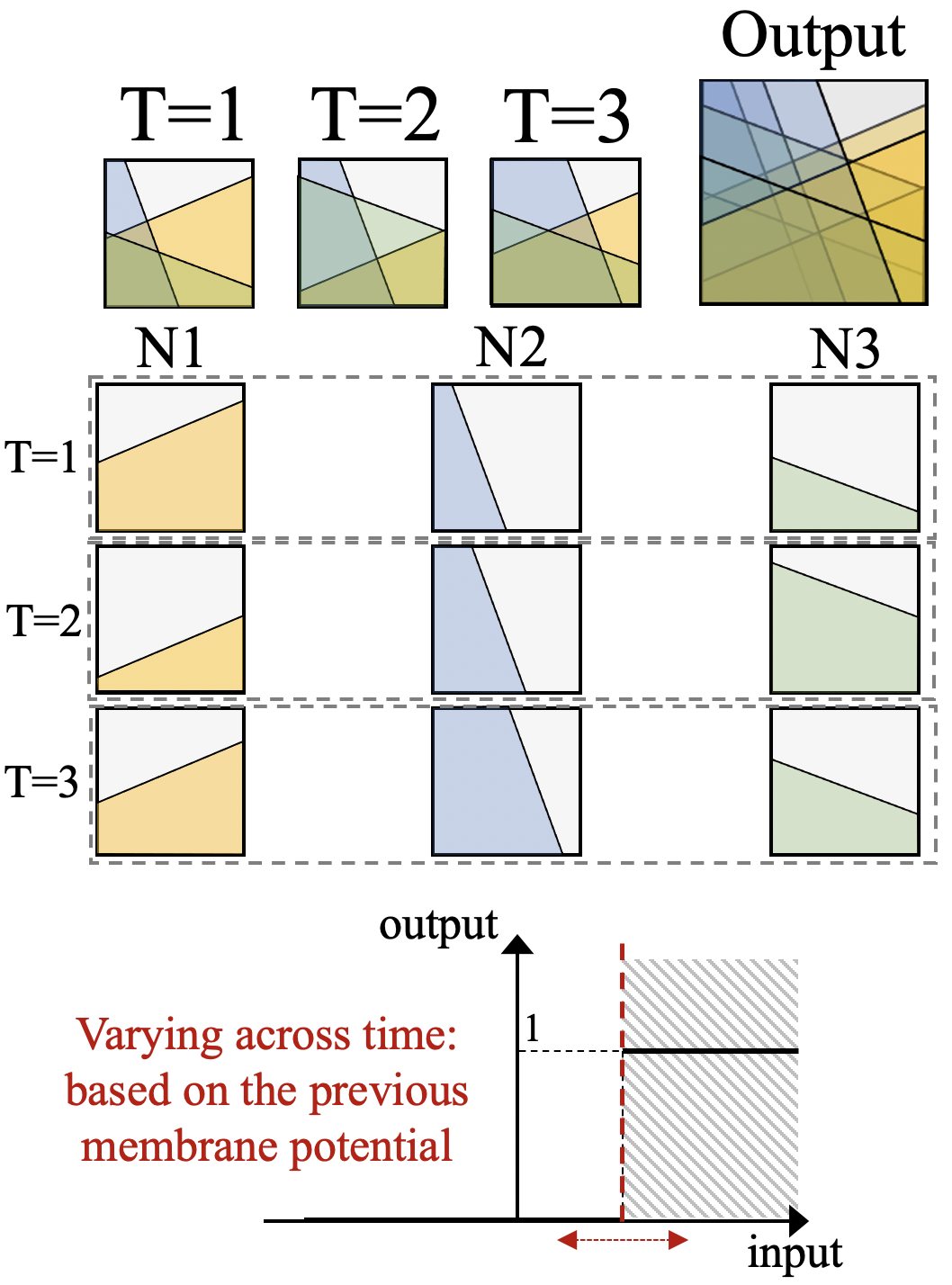} &
\\
{(a) ReLU } & { (b)  LIF} \\
\end{tabular}
\end{center}
\caption{Illustration of the concept of linear regions from ReLU and LIF neurons.
Each ReLU (or LIF neuron) divides the two-dimension input space into active and inactive regions. 
}
\label{fig:method:lenear_concept}
\end{figure}
\end{minipage}
%
%
\hfill
\begin{minipage}{0.49\textwidth}
\centering
\begin{figure}[H]
\begin{center}
\def\arraystretch{0.5}
\begin{tabular}{@{}c@{\hskip 0.03\linewidth}c@{}c}
\includegraphics[width=0.50\linewidth]{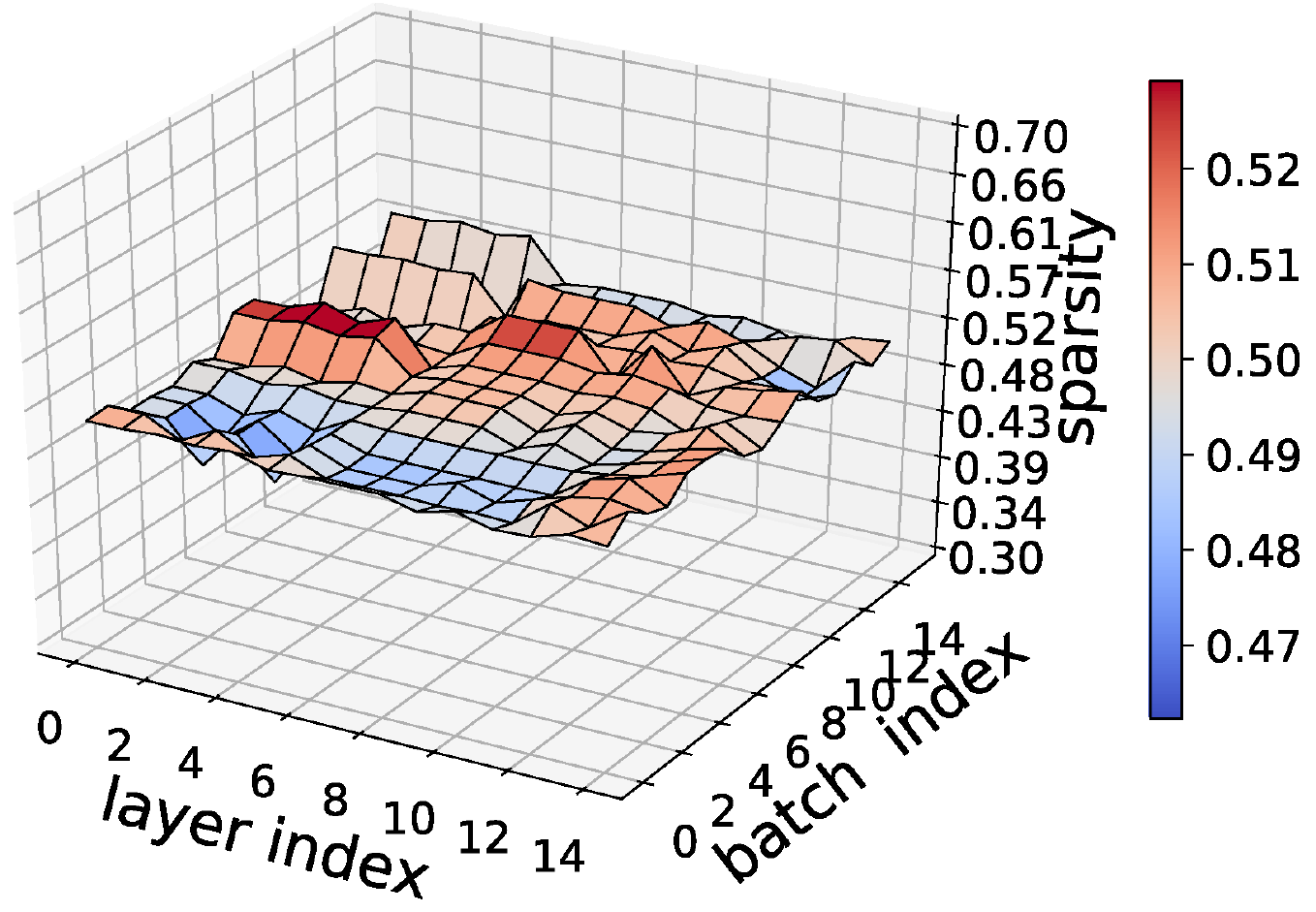} &
\includegraphics[width=0.50\linewidth]{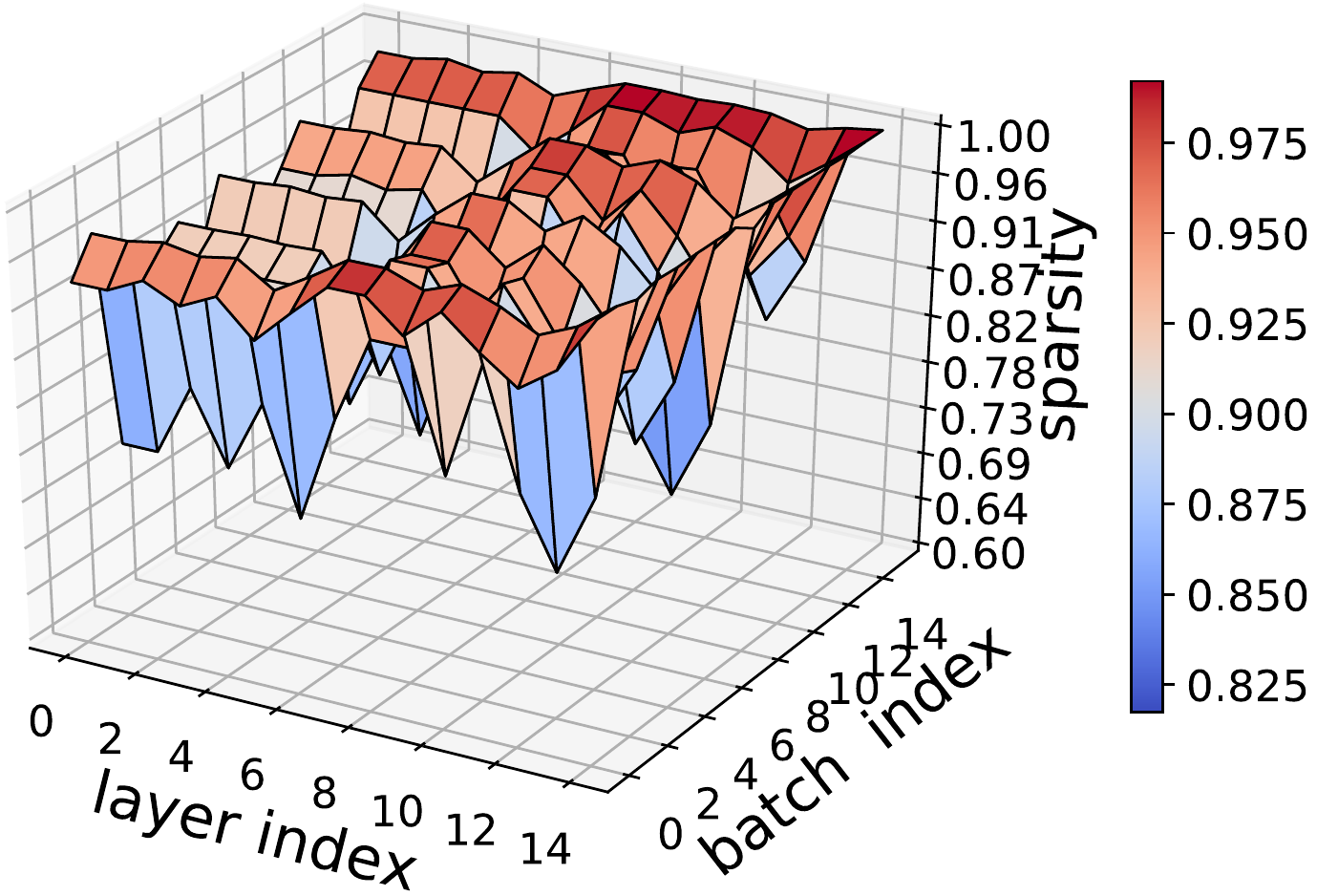} 
\\
{ (a) ReLU } & { (b)  LIF (t=1) }\\
\includegraphics[width=0.50\linewidth]{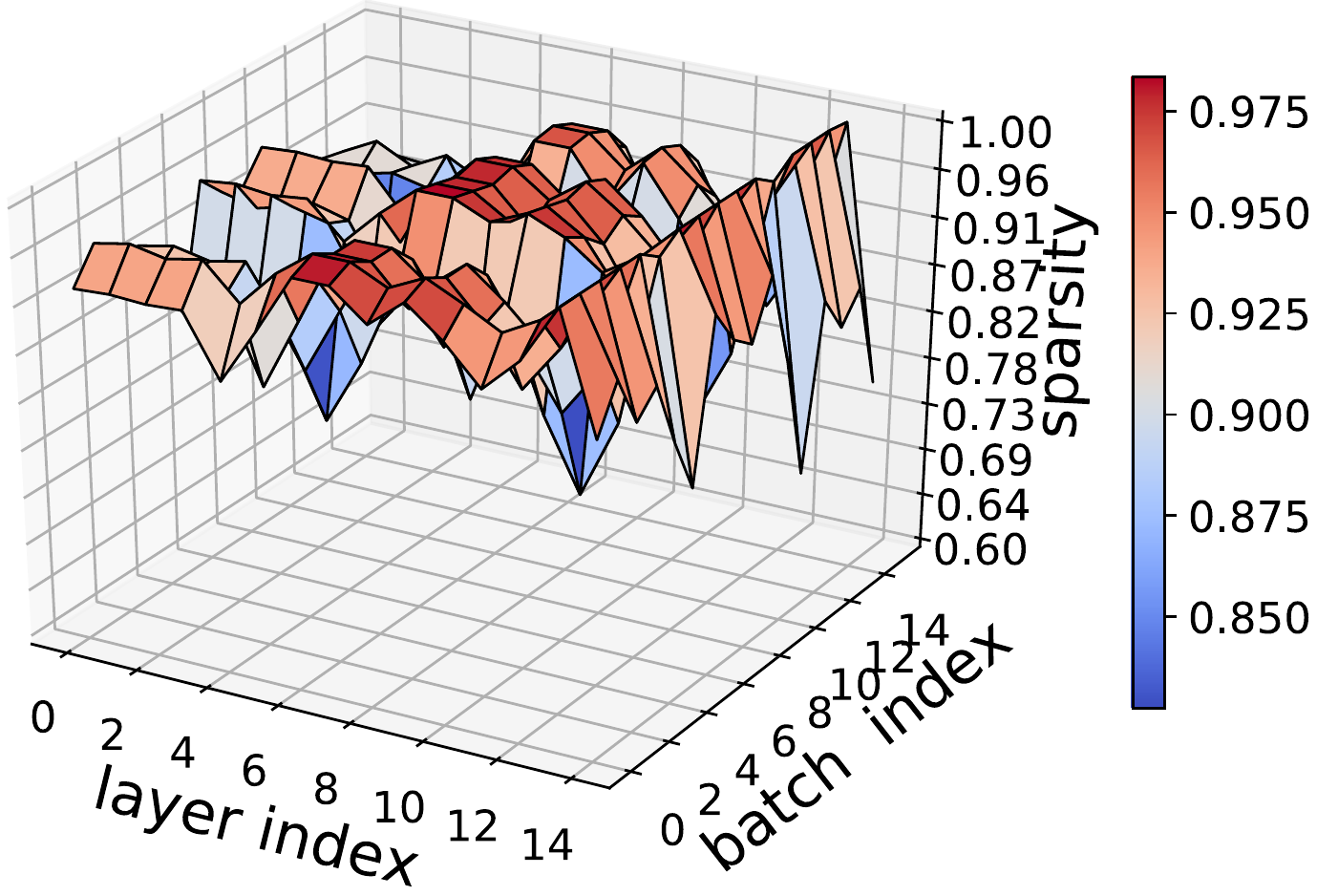} &
\includegraphics[width=0.50\linewidth]{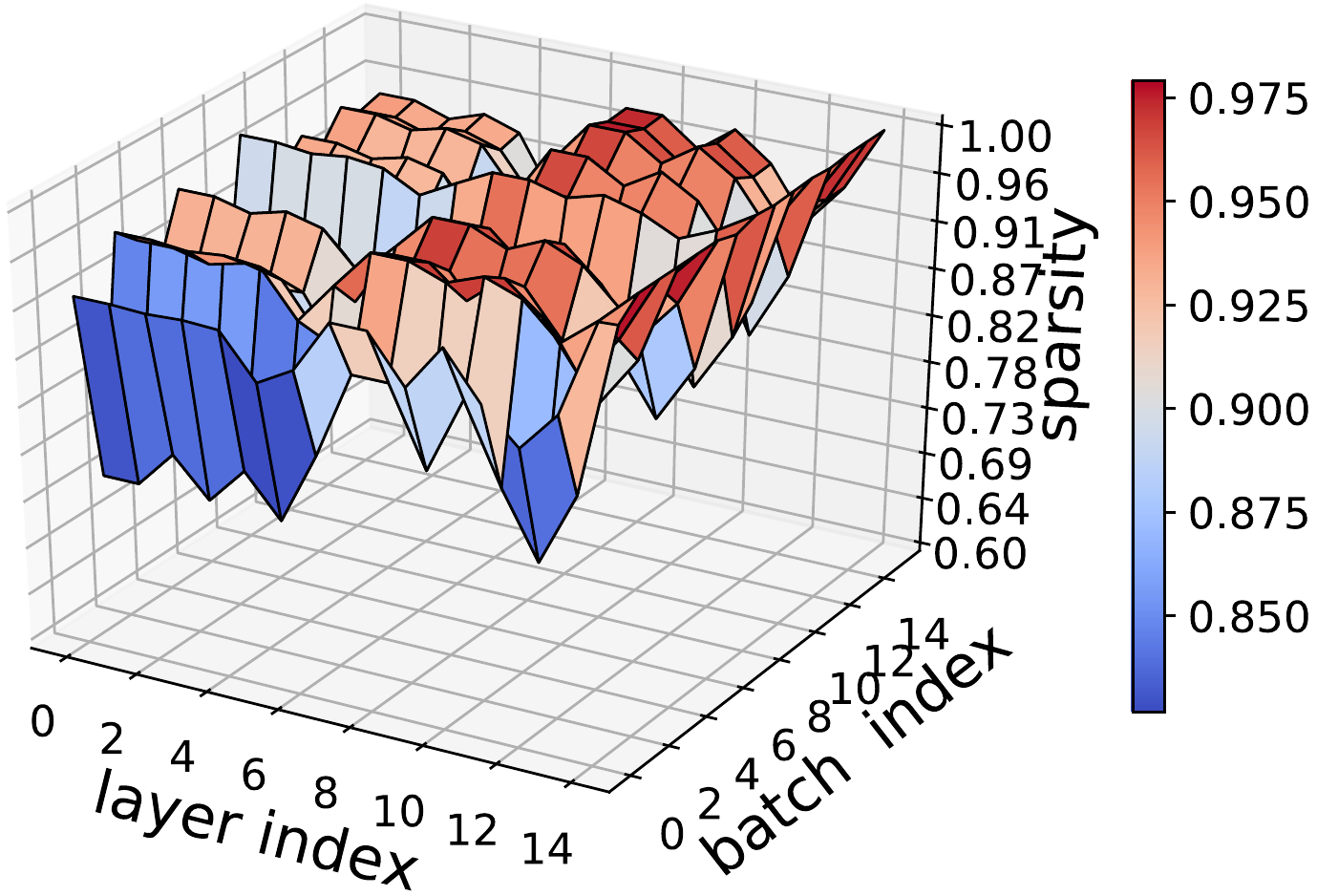} 
\\
{(c) LIF (t=3) } & {(d) LIF (t=5) }\\
\end{tabular}
\end{center}
\caption{Sparsity variance of activation pattern across different layers and different samples in a mini-batch. LIF neuron shows higher variance of activation pattern compared to ReLU neuron.
}
 \label{fig:method:sparsity}
\end{figure}
\end{minipage}
\end{minipage}
%
\end{figure}

\noindent
For instance, ReLU and Hard Tanh have neuron transition at $0$ and $\{-1, 1\}$, respectively \cite{raghu2017expressive}.
Fig. \ref{fig:method:lenear_concept}(a) also shows the simple example with three ReLU neurons. 
The input space is divided into two regions by a single ReLU neuron. 
By composing ReLU neurons, the input space is partitioned into multiple regions where each region represents a different linear function.

According to Definition 1, a LIF neuron can be regarded as a piecewise linear function.
For each timestep, the LIF neuron transfers $0$ if the membrane potential is lower than a firing threshold, otherwise it generates $1$ (\ie, spike). 
Thus, neuron transition occurs when a given input generates an output spike. We illustrate the transfer function of an LIF neuron in Fig. \ref{fig:method:lenear_concept}(b).
Different from ReLU neuron, the output of LIF neuron is not solely dependent on the input. 
As we shown in Eq. \ref{eq:LIF}, the output of LIF neuron is based on the current input as well as the previous membrane potential.
Therefore, the neuron transition point can be changing across time.
For example, suppose that the firing threshold is $1$ and the membrane potential from the previous timestep is $0.3$. In this case, neuron transition happens at $input = 0.7$. 
After the neuron fires, the membrane potential is reset to $0$, where, the neuron transition point becomes $1$.
With this time-varying transfer function, the linear region of SNNs becomes more diverse.

\subsection{Sparsity-Aware Hamming Distance}

In NASWOT \cite{mellor2021neural}, Hamming Distance (HD) is a key metric to compare the binary activation pattern $\cB_i, \cB_j$ between two different mini-batch samples $i, j$. 
However, standard HD gives inaccurate distance measurement for SNNs due to the large sparsity variance of binary activation pattern $\cB$ of LIF.
Here, the term ``sparsity" denotes the percentage of $0$ in binary activation pattern $\cB$ from one layer at a given timestep $t$.
Note, the definition of ``sparsity" here is slightly different from the previous works which defines ``sparsity" from the activation across all timesteps.

\noindent \textbf{Observation on the sparsity of activation pattern.} A ReLU neuron provides a binary activation pattern with about $50\%$ sparsity from Gaussian/Uniform weight initialization \cite{he2015delving}, which is similar across all data samples.
On the other hand, a LIF neuron shows a large variation of sparsity across different data samples since the activation pattern is based on the previous membrane potential which is different in each sample.
In Fig. \ref{fig:method:sparsity}, we visualize the sparsity of binary activation pattern with $16$ mini-batch samples.
The results demonstrate that LIF neuron causes a large sparsity variation across different samples.

\begin{figure}[t]
\noindent\begin{minipage}{\textwidth}
\begin{minipage}{0.6\textwidth}
\centering
%
\begin{figure}[H]
\begin{center}
\def\arraystretch{0.5}
\begin{tabular}{@{}c@{\hskip 0.01\linewidth}c@{}c}
\includegraphics[width=0.49\linewidth]{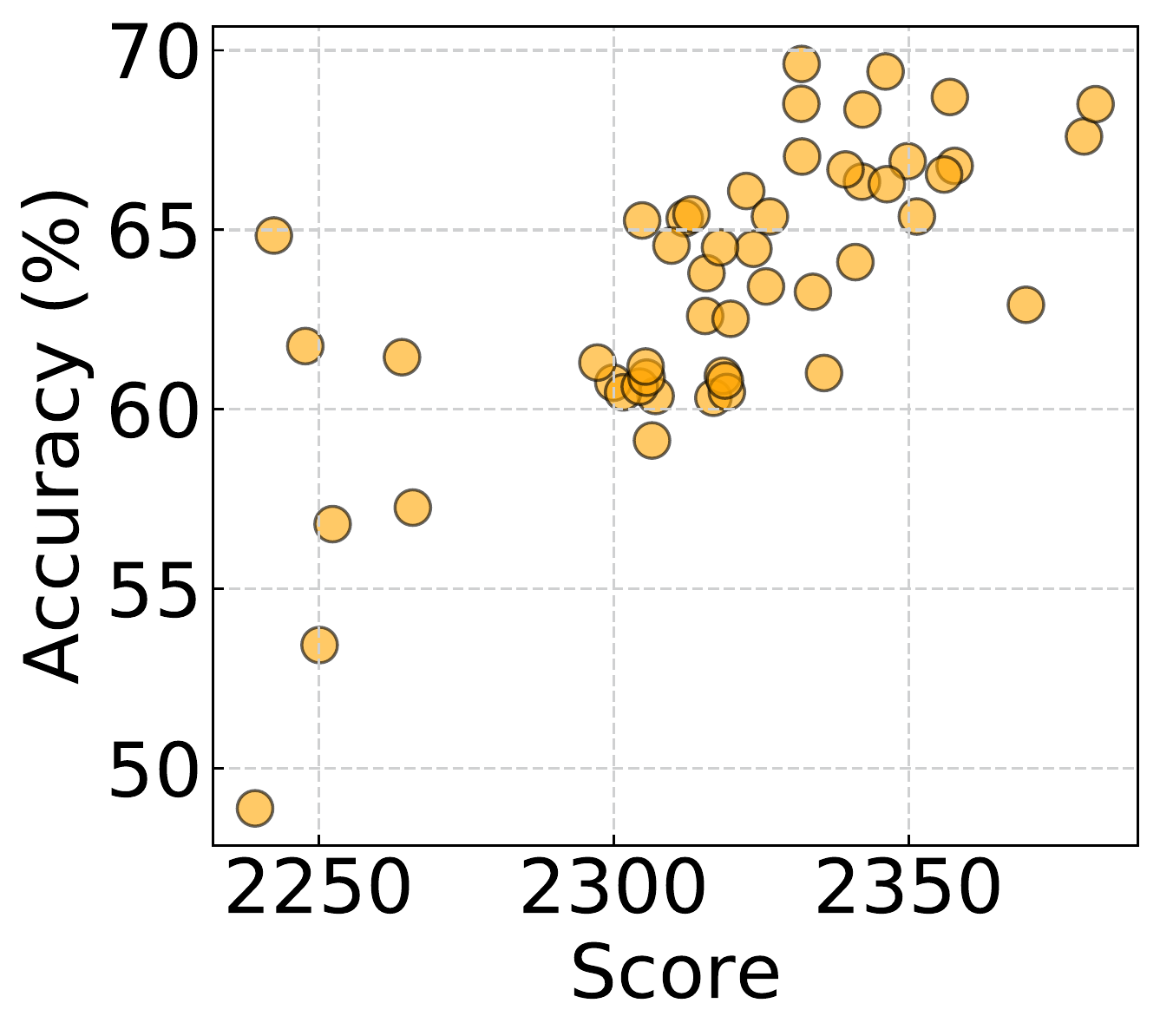}&
\includegraphics[width=0.52\linewidth]{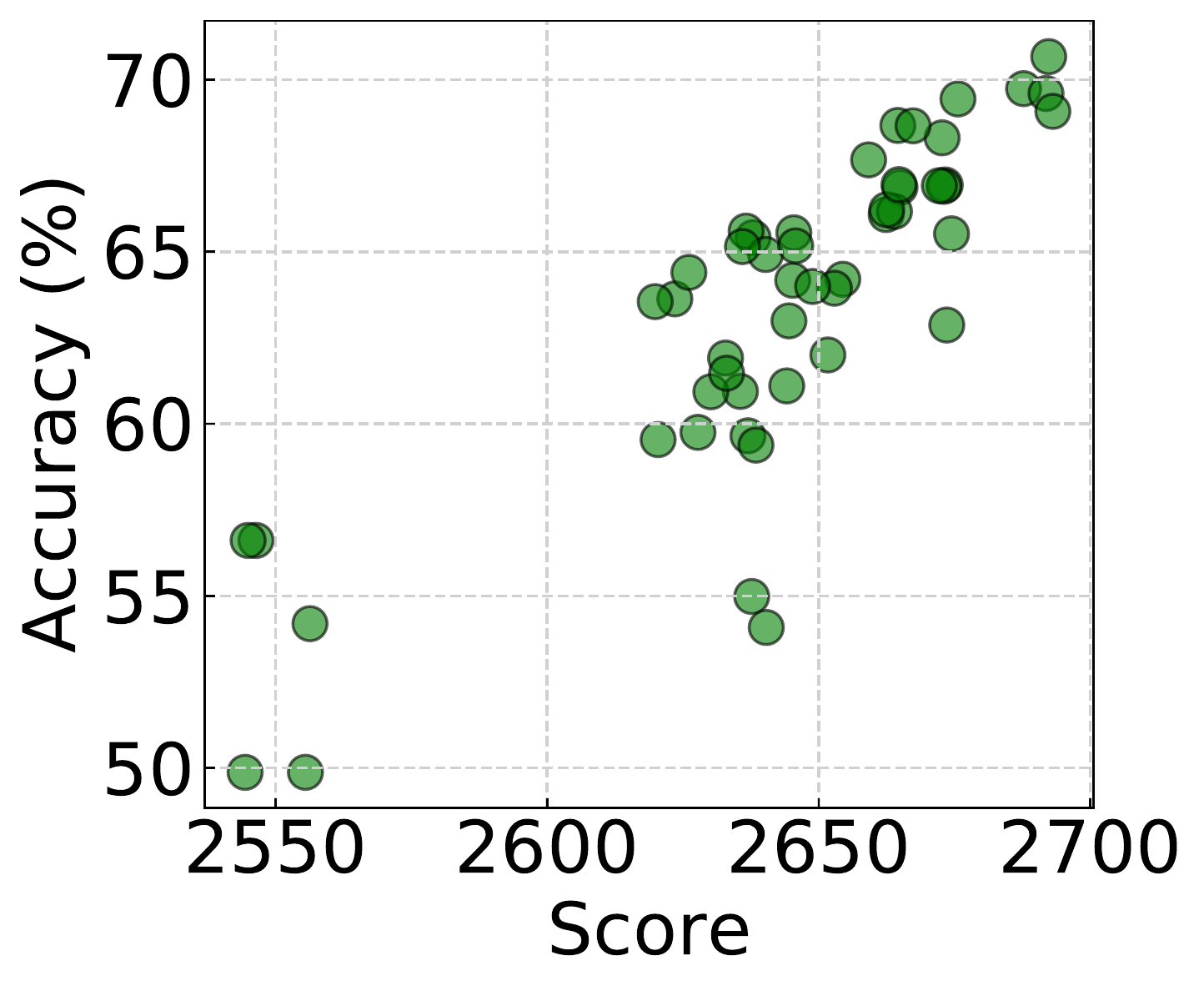}
\\
{ (a) HD } & { (b) SAHD }
\\
\end{tabular}
\end{center}
\caption{ 
Accuracy with respect to architecture score.
We randomly select 50 architectures
from search space.
We show Kendall's $\tau$ correlation  for quantitative comparison (HD: $\tau$=0.519, SAHD: $\tau$= 0.646).
}
\label{fig:method:kendall}
\end{figure}
\end{minipage}
%
%
\hfill
\begin{minipage}{0.38\textwidth}
\centering
\begin{figure}[H]
\begin{center}
         \includegraphics[width=1.0\textwidth]{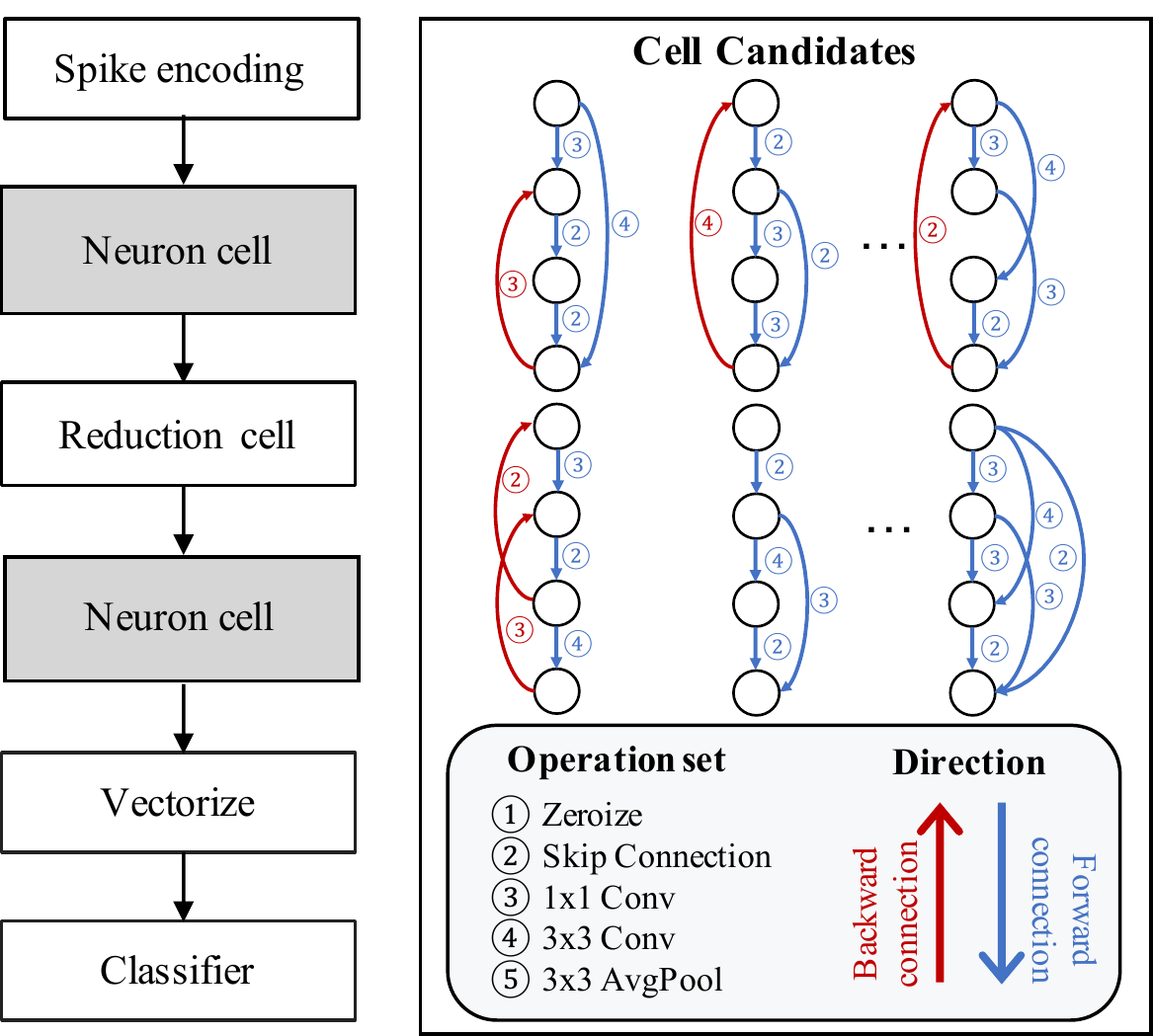}
\end{center}
 \caption{
Illustration of cell-based neural architecture search. 
 }
     \label{fig:method:cellbased-nas}

\end{figure}
\end{minipage}
\end{minipage}
%
\end{figure}

\noindent \textbf{A problem due to large sparsity variation.} This large sparsity variation induces different scales of HD.
To explain this, for the data sample $i$, we model the distribution of LIF neuron output (at each timestep) as i.i.d. Bernoulli distribution where the probability of observing $1$ is $1-r_{i}^{l}$:
\begin{equation}
    o_{i}^{l} \sim \textup{Bern}(1-r_{i}^{l}).
    \label{eq: bern}
\end{equation}
Here, $r_{i}^{l}$ is sparsity of binary activation pattern at layer $l$.
Then, the probability of an activation difference (at the same neuron position) between two data samples $i$, $j$ can be represented as:
\begin{equation}
Pr(|o^l_{i}-o^l_{j}|=1) =  \textup{Bern}(r_{i}^{l}(1-r_{j}^{l}) + (1-r_{i}^{l})r_{j}^{l}).
\label{eq: pr_diff1}
\end{equation}
Considering that each element of binary activation patterns $\cB^l \in \R^{N^{l}_A}$ is sampled from  Bernoulli distribution (Eq. \ref{eq: bern}), where $N^{l}_A$ denotes the number of neurons at layer $l$. 
Then, the expectation of HD (\ie, $d_H(\cB^l_i, \cB^l_j)$) can be formulated as:
\begin{equation}
\E[d_H(\cB^l_i, \cB^l_j)] = N^l_A \E[ Pr(|o^l_{i}-o^l_{j}|=1)] = N^l_A \{r_{i}^{l}(1-r_{j}^{l}) + (1-r_{i}^{l})r_{j}^{l}\}, 
\label{eq: exp_dh}
\end{equation}
Note, all quantities in Eq. \ref{eq: bern}, Eq. \ref{eq: pr_diff1}, and  Eq. \ref{eq: exp_dh} are evaluated per timestep, and we average them across timesteps.

As we can observe in Eq. \ref{eq: exp_dh}, the expectation of HD is the function of sparsity $r_{i}^{l}$ and $r_{j}^{l}$.
Therefore, HD will provide an inaccurate distance measure for SNN where sparsity $r^{l}$ has a large variation across data samples (Fig. \ref{fig:method:sparsity}).
For example, HD is likely to be small if two activations are in extreme cases, highly-sparse ($r\rightarrow1$) or highly-dense ($r \rightarrow 0$).
On the other hand, HD is likely to be high if two activations are in a moderate range ($r \approx 0.5$).
Thus, based on the sparsity of two activations, HD has a different contribution to the final score $s$ (Eq. \ref{eq:score-wot}); the ideal case is when all HD have the same contribution.

\noindent \textbf{The proposed solution.} 
To address this problem, we propose Sparsity-Aware Hamming Distance (SAHD) where Hamming Distance is normalized based on the sparsity of two binary activation patterns.
This can be simply done by normalizing the expectation of HD value to a constant $\alpha$:
\begin{equation}
 d_{SAH}(\cB^l_i, \cB^l_j) =  \frac{\alpha}{N^l_A\{r_{i}^{l}(1-r_{j}^{l}) + (1-r_{i}^{l})r_{j}^{l}\}} d_H(\cB^l_i, \cB^l_j).
 \label{eq:SAHD}
\end{equation}
We compute the global SAHD score by accumulating layer-wise SAHD across all layers, \ie, $d_{SAH}(\cB_i, \cB_j)=\sum_l d_{SAH}(\cB^l_i, \cB^l_j)$.
Instead of HD, we use SAHD for computing the kernel matrix (Eq. \ref{eq:KH-wot}) at each timestep. After that, we sum all kernel matrices to compute the final score using Eq. \ref{eq:score-wot}.
In Fig. \ref{fig:method:kendall}, we compare the correlation between 
architecture score and post-training accuracy for HD and SAHD.
The results demonstrate that the proposed SAHD has a higher Kendall's $\tau$ value which implies it is a more accurate metric for architecture selection.

\subsection{Searching Forward and Backward Connections}

Cell-based approach \cite{zoph2018learning,real2019regularized,pham2018efficient,shu2019understanding,liu2018darts,xie2018snas} is widely used in NAS research.
These methods usually search for the connection topology as well as the corresponding operation for each connection. 
Then, multiple generated cell architectures construct the whole network.
In our search algorithm, we also investigate cell-based architectures.
Fig. \ref{fig:method:cellbased-nas} shows the macro skeleton of our SNN architecture. 
The first block is the spike encoding layer which directly converts a float value image into spikes like previous works \cite{wu2019direct,zheng2020going,zhang2020temporal}.
The main body of the skeleton consists of two searched neuron cells and one reduction cell.
The reduction cell includes one convolution layer and 2-by-2 Average pooling with stride 2.
Finally, a linear classifier is used for prediction.

\noindent\textbf{Cell Search Strategy.}
Our cell search space is identical to NAS-Bench-201 \cite{dong2020bench} (except for backward connections) where each cell includes $V=4$ nodes with multiple connections  sampled from operation set $O =$ \textit{\{zeroize, skip connection, 1-by-1 convolution, 3-by-3 convolution, 3-by-3 average pooling\}} (see Fig. \ref{fig:method:cellbased-nas}).
Each node contains the sum of all incoming feature maps from edge operation. 
However, different from \cite{dong2020bench}, we search backward connections in addition to forward connections.
In backward operation, we add transformed node feature of $l$-th layer at timestep $t-1$ to the node of $l'$-th ($l'<l$) layer at timestep $t$.
The backward connections also have the same operation set search space $O$  as forward connections.
In Fig. \ref{fig:method:cellbased-nas}, we show examples of cell candidates. In the predefined search space, we select the optimal spiking neuron cell. Reduction cell downsamples the spatial size of the feature map. 
We do not illustrate Zeroize operation for simplicity.
The forward connections and backward connections can be combined seamlessly.
Surprisingly, adding backward connections improves the accuracy of SNNs especially on complex datasets such as CIFAR100 and Tiny-ImageNet.
To train the searched SNNs, we use surrogate gradient training \cite{neftci2019surrogate,wu2018spatio,wu2019direct} (see Supplementary C for details).

\section{Experiments}

\subsection{Implementation Details}

\noindent \textbf{Dataset.} We evaluate our method on CIFAR10 \cite{krizhevsky2009learning}, CIFAR100 \cite{krizhevsky2009learning}, TinyImageNet \cite{deng2009imagenet}. The details of datasets can be found in Supplementary.

\noindent \textbf{Hyperparameters.} Our implementation is based on PyTorch \cite{paszke2017automatic}. We train the networks with standard SGD with momentum 0.9,  weight decay 0.0005
and also apply random crop and horizontal flip to input images. We set batch size for training as 64.
The base learning rate is set to 0.2, 0.1, 0.1 for CIFAR10, CIFAR100, TinyImageNet, respectively. 
We use cosine learning rate scheduling~\cite{loshchilov2016sgdr}. Here, we set the total number of epochs to 300, 300, 200, for CIFAR10, CIFAR100, TinyImageNet, respectively. 
We set $\tau_m$ in Eq.~\ref{eq:LIF} to $\frac{4}{3}$.
We set $\alpha$ in Eq.~\ref{eq:SAHD} to $0.5N_{A}^l$ to get similar sparsity scale in LIF neuron as a ReLU neuron.
Also, we search 5000 architecture candidates from search space (We observe the accuracy saturates after 5000 samples, shown in Supplementary E).
We use SpikingJelly \cite{SpikingJelly} package for implementing an LIF neuron.

\begin{table*}[t]
    \addtolength{\tabcolsep}{2.5pt}
    \centering
    \caption{Classification Accuracy (\%)  on CIFAR10, CIFAR100, and TinyImageNet.}
    \resizebox{0.84\textwidth}{!}{%
    \begin{tabular}{lccccc}
        \toprule
        & \:\:\:\textrm{Dataset}\:\:\: 
        & \:\:\:\textrm{Training Method}\:\:\: & \:\:\:\textrm{Architecture}\:\:\: & \:\:\:\textrm{Timesteps}\:\:\: & \:\:\:\:\:\textrm{Accuracy(\%)}\:\:\: \\
         \midrule
        Wu \textit{et al.} \cite{wu2019direct} & CIFAR10 & Surrogate Gradient & 5Conv, 2Linear  & 12 & 90.53  \\
        Wu \textit{et al.} \cite{wu2019tandem} & CIFAR10 & Tandem Learning  & 5Conv, 2Linear  & 8 & 89.04 \\
        Rathi \textit{et al.} \cite{rathi2020enabling}& CIFAR10 & Hybrid & VGG9 & 100 & 90.50 \\ 
        Han \textit{et al.} \cite{han2020rmp}& CIFAR10 & ANN-SNN Conversion & VGG16 & 2048 & 93.63  \\
        Kundu \textit{et al.} \cite{kundu2021spike}& CIFAR10 & Hybrid & VGG16 & 100 & 91.29\\
        Zheng \textit{et al.} \cite{zheng2020going}& CIFAR10 & Surrogate Gradient & ResNet19  & 6 &  93.16  \\
        Deng \textit{et al.} \cite{deng2021optimal}& CIFAR10 & ANN-SNN Conversion & ResNet20 & 16 & 92.42 \\
        Li \textit{et al.} \cite{li2021free}& CIFAR10 & ANN-SNN Conversion & VGG16 & 32 & 93.00 \\
        Fang \textit{et al.} \cite{fang2021incorporating}& CIFAR10 & Surrogate Gradient & 6Conv, 2Linear & 8 &  93.50  \\
        Rathi \textit{et al.} \cite{rathi2021diet}& CIFAR10 & Hybrid & VGG16 & 5 & 92.70 \\
        \cellcolor{blue!10}SNASNet-Fw (ours) & \cellcolor{blue!10}CIFAR10 & \cellcolor{blue!10} Surrogate Gradient & \cellcolor{blue!10} Searched Architecture & \cellcolor{blue!10}5 &  \cellcolor{blue!10} 93.12 $\pm$ 0.42  \\
        \cellcolor{blue!10}SNASNet-Fw (ours) & \cellcolor{blue!10}CIFAR10 & \cellcolor{blue!10} Surrogate Gradient & \cellcolor{blue!10} Searched Architecture & \cellcolor{blue!10}8 &  \cellcolor{blue!10} 93.64 $\pm$ 0.35  \\
        \cellcolor{red!10}SNASNet-Bw (ours) &  \cellcolor{red!10}CIFAR10 & \cellcolor{red!10} Surrogate Gradient & \cellcolor{red!10} Searched Architecture &\cellcolor{red!10}5 &  \cellcolor{red!10} 93.73 $\pm$ 0.32  \\
        \cellcolor{red!10}SNASNet-Bw (ours) &  \cellcolor{red!10}CIFAR10 & \cellcolor{red!10} Surrogate Gradient & \cellcolor{red!10} Searched Architecture &\cellcolor{red!10}8 &  \cellcolor{red!10} 94.12 $\pm$ 0.25  \\
        \midrule
        Lu and Sengupta \cite{lu2020exploring}& CIFAR100 & ANN-SNN Conversion & VGG15 & 62 & 63.20  \\
        Park \textit{et al.} \cite{park2020t2fsnn}& CIFAR100 & TTFS & VGG15 & 680 & 68.80  \\
        Rathi \textit{et al.}  \cite{rathi2020enabling} & CIFAR100 & Hybrid & VGG16 & 125 & 67.80 \\ 
        Han \textit{et al.} \cite{han2020rmp}& CIFAR100 & ANN-SNN Conversion & VGG16 & 2048 & 70.90  \\
        Garg \textit{et al.} \cite{garg2021dct}& CIFAR100 & DCT & VGG9 & 48 & 68.30 \\
        Kundu \textit{et al.} \cite{kundu2021spike}& CIFAR100 & Hybrid & VGG11 & 120 & 64.98\\
        Deng \textit{et al.} \cite{deng2021optimal}& CIFAR100 & ANN-SNN Conversion & ResNet20 & 32 & 68.40 \\
        Li \textit{et al.} \cite{li2021free}& CIFAR100 & ANN-SNN Conversion & ResNet20 & 16 & 72.33 \\
        Rathi \textit{et al.} \cite{rathi2021diet}& CIFAR100 & Hybrid & VGG16 & 5 & 69.67 \\
        \cellcolor{blue!10}SNASNet-Fw (ours) & \cellcolor{blue!10}CIFAR100 & \cellcolor{blue!10}Surrogate Gradient & \cellcolor{blue!10}Searched Architecture & \cellcolor{blue!10}5 & \cellcolor{blue!10} 70.06 $\pm$ 0.45  \\
        \cellcolor{red!10}SNASNet-Bw (ours) & \cellcolor{red!10}CIFAR100 & \cellcolor{red!10}Surrogate Gradient & \cellcolor{red!10}Searched Architecture & \cellcolor{red!10}5 &  \cellcolor{red!10} 73.04 $\pm$ 0.36  \\
        \midrule
         Sengupta \textit{et al.} \cite{sengupta2019going}& TinyImageNet & ANN-SNN Conversion & VGG11 & 2500 & 54.20\\
         Kundu \textit{et al.} \cite{kundu2021spike}& TinyImageNet & Hybrid & VGG16 & 150 & 51.92\\
         Garg \textit{et al.} \cite{garg2021dct}& TinyImageNet & DCT & VGG13 & 125 & 52.43 \\
        \cellcolor{blue!10}SNASNet-Fw (ours) & \cellcolor{blue!10}TinyImageNet & \cellcolor{blue!10}Surrogate Gradient & \cellcolor{blue!10} Searched Architecture &\cellcolor{blue!10}5 & \cellcolor{blue!10} 52.81 $\pm$ 0.56  \\
        \cellcolor{red!10}SNASNet-Bw (ours) & \cellcolor{red!10}TinyImageNet & \cellcolor{red!10}Surrogate Gradient &\cellcolor{red!10} Searched Architecture & \cellcolor{red!10}5 &  \cellcolor{red!10} 54.60 $\pm$ 0.48 \\
        \bottomrule
        \\
    \end{tabular}%
    }
    \label{table:performance}
\end{table*}

\noindent \textbf{Architectures.}
Here, we provide details for architectures in Fig. \ref{fig:method:cellbased-nas}. Note, we do not allow two nodes to have both forward and backward connections to ensure training convergence and stability. For the spike encoding layer, we use direct coding \cite{wu2019direct,zheng2020going,zhang2020temporal} where we pass the input image for $T$ time-steps through the first convolution layer which generates spikes. 
The first neuron cell has \textit{C}-channel input and \textit{C}-channel output.
Reduction cell consists of \textit{Conv(\textit{C}, 2\textit{C})-BN(2\textit{C})-LIF} followed by \textit{AvgPool(2)}.
The second neuron cell has 2\textit{C}-channel input and 2\textit{C}-channel output. Note, the structures of the first neuron cell and second neuron cell are identical.
We set \textit{C} to 256, 128, 128 for CIFAR10, CIFAR100, TinyImagNet, respectively.
For vectorize block, we first apply \textit{AvgPool(2)} to the input feature and vectorize the output.
Finally, the classifier consists of \textit{Dropout(0.5)-FC(1024)-Voting layer}, where a voting layer is used to improve the robustness of classification \cite{fang2021incorporating}.

\begin{figure}[t]
\begin{center}
\def\arraystretch{0.5}
\begin{tabular}{@{\hskip -0.004\linewidth}c@{\hskip 0.02\linewidth}c@{\hskip 0.02\linewidth}c@{\hskip 0.02\linewidth}c@{\hskip 0.02\linewidth}c@{\hskip 0.02\linewidth}c@{\hskip 0.02\linewidth}}
\includegraphics[width=0.09\linewidth]{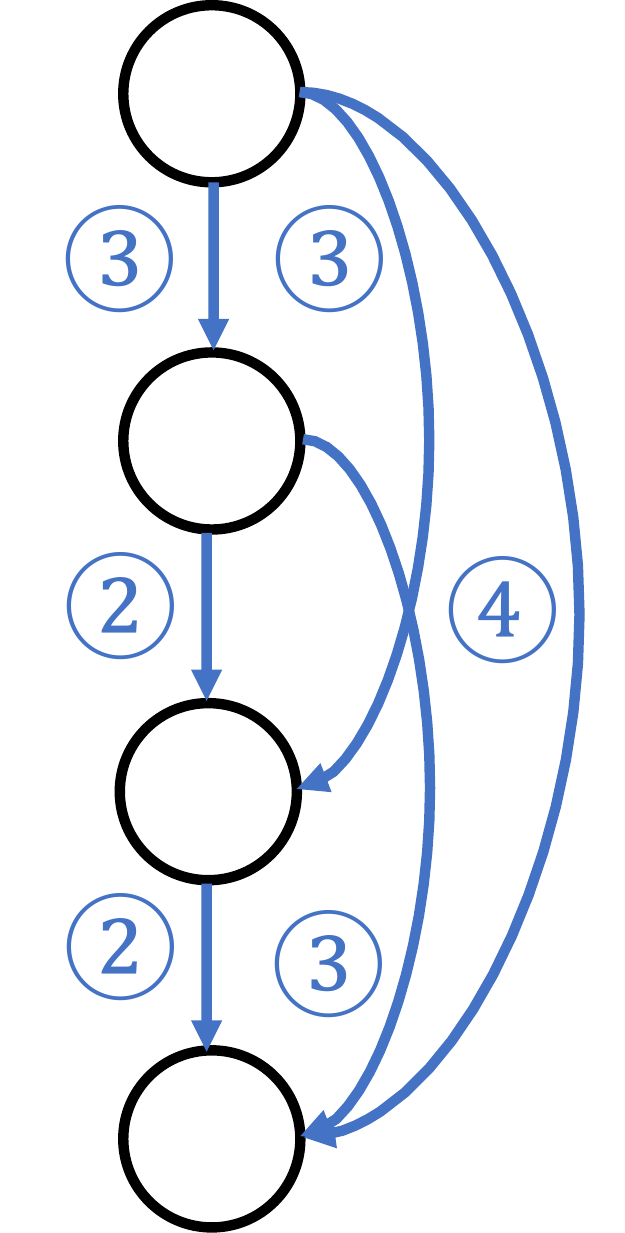} &
\includegraphics[width=0.09\linewidth]{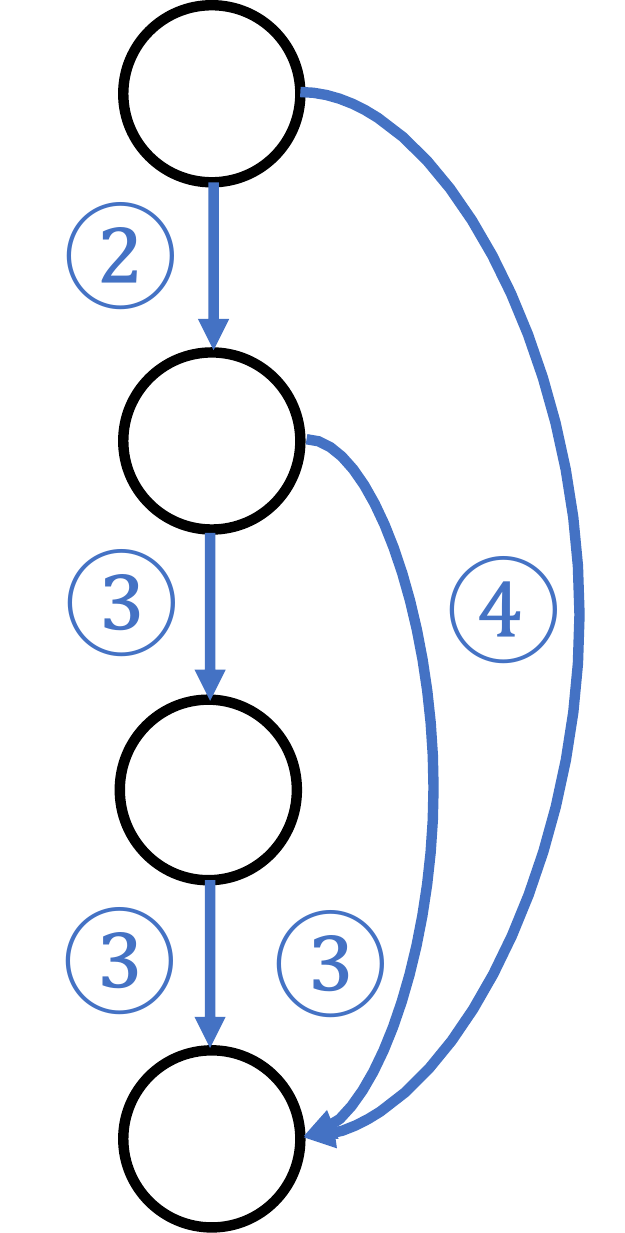} &
\includegraphics[width=0.09\linewidth]{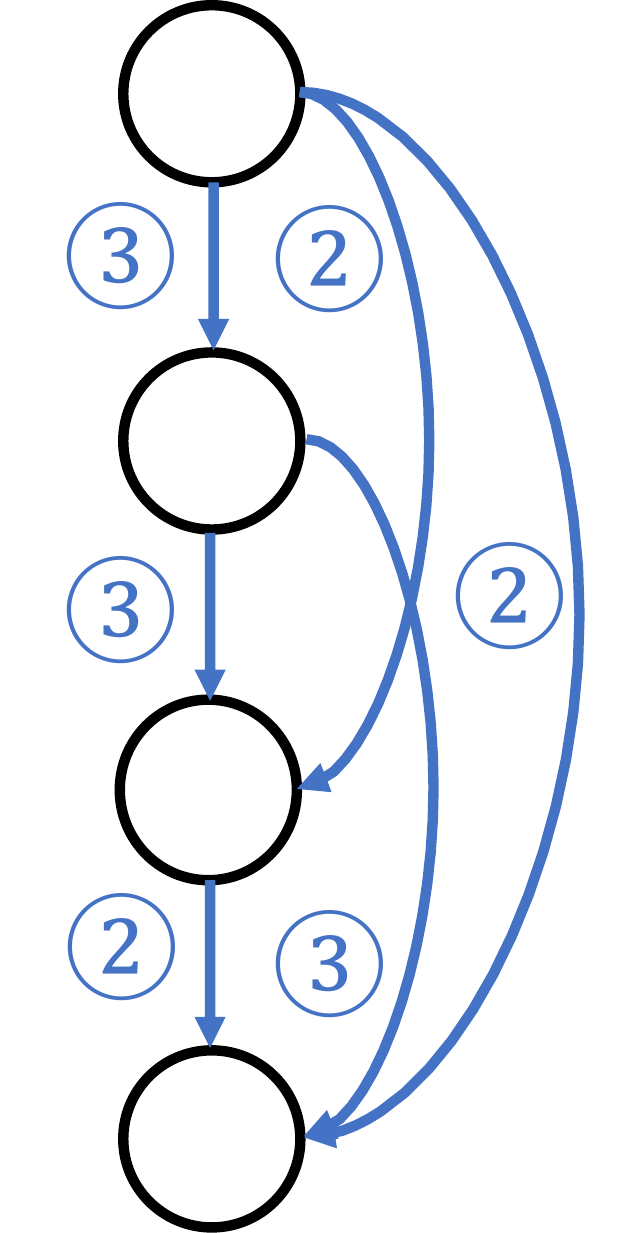} &
\includegraphics[width=0.103\linewidth]{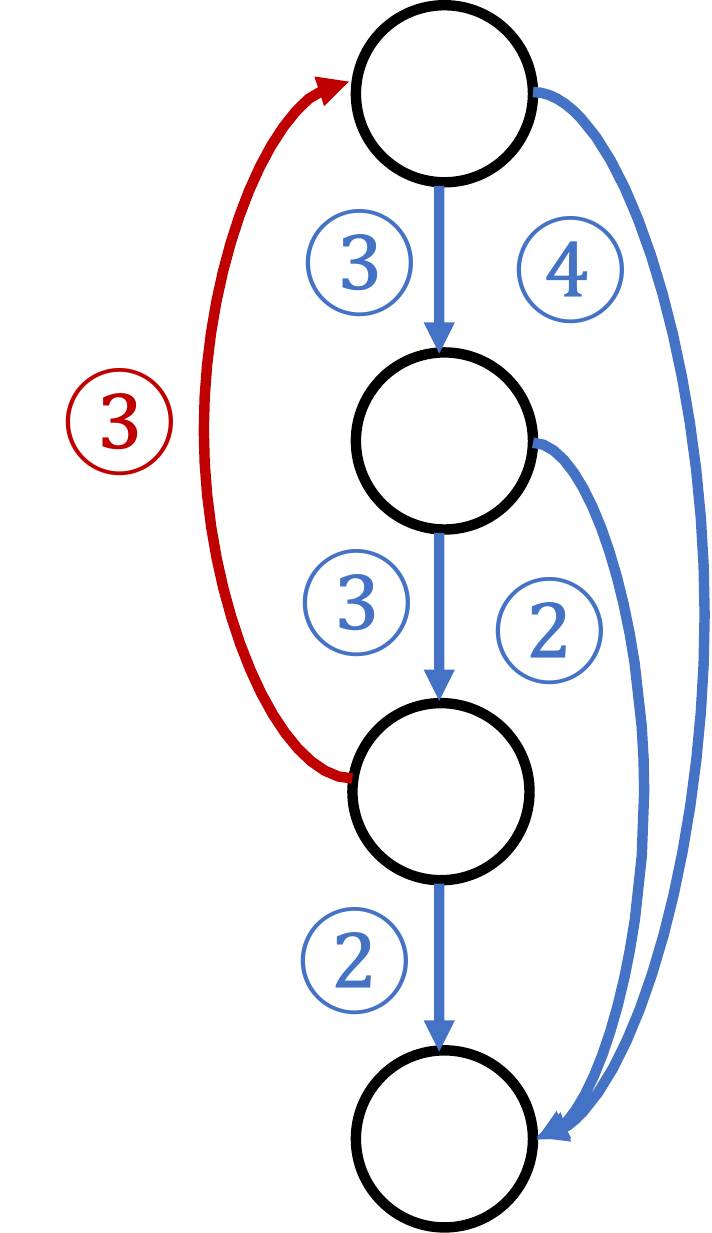} &
\includegraphics[width=0.105\linewidth]{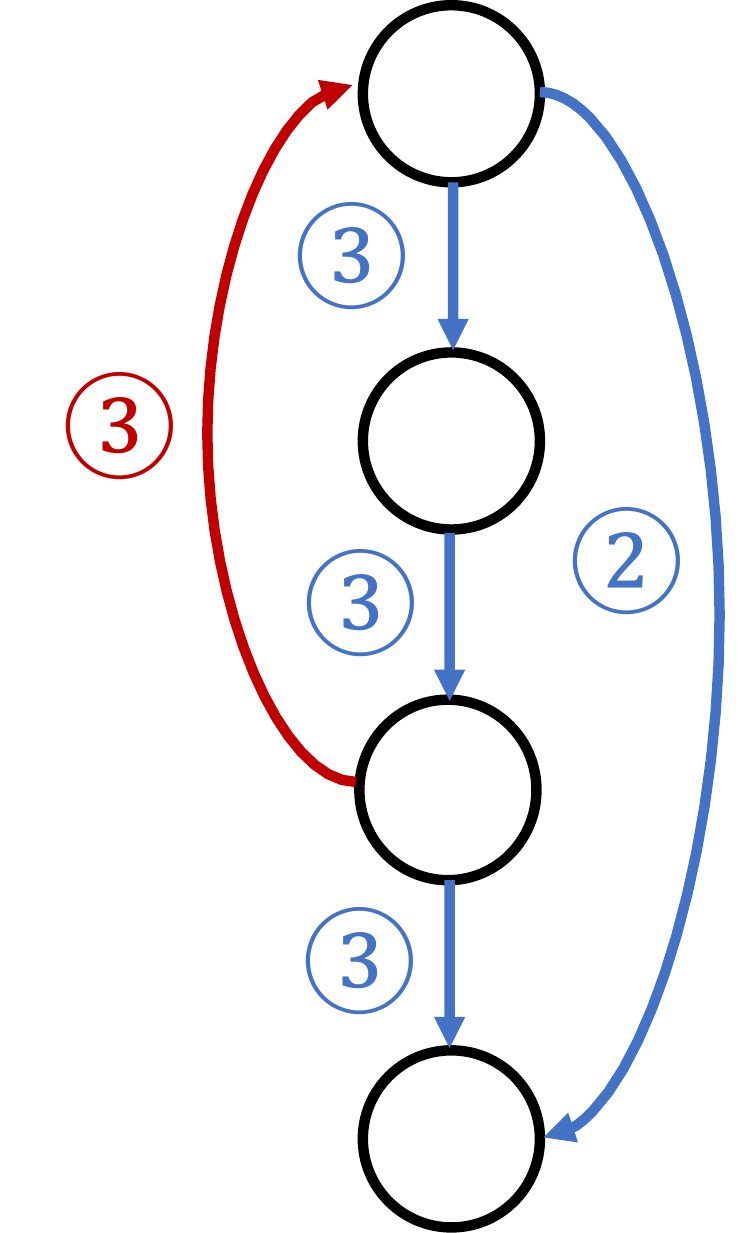} &
\includegraphics[width=0.103\linewidth]{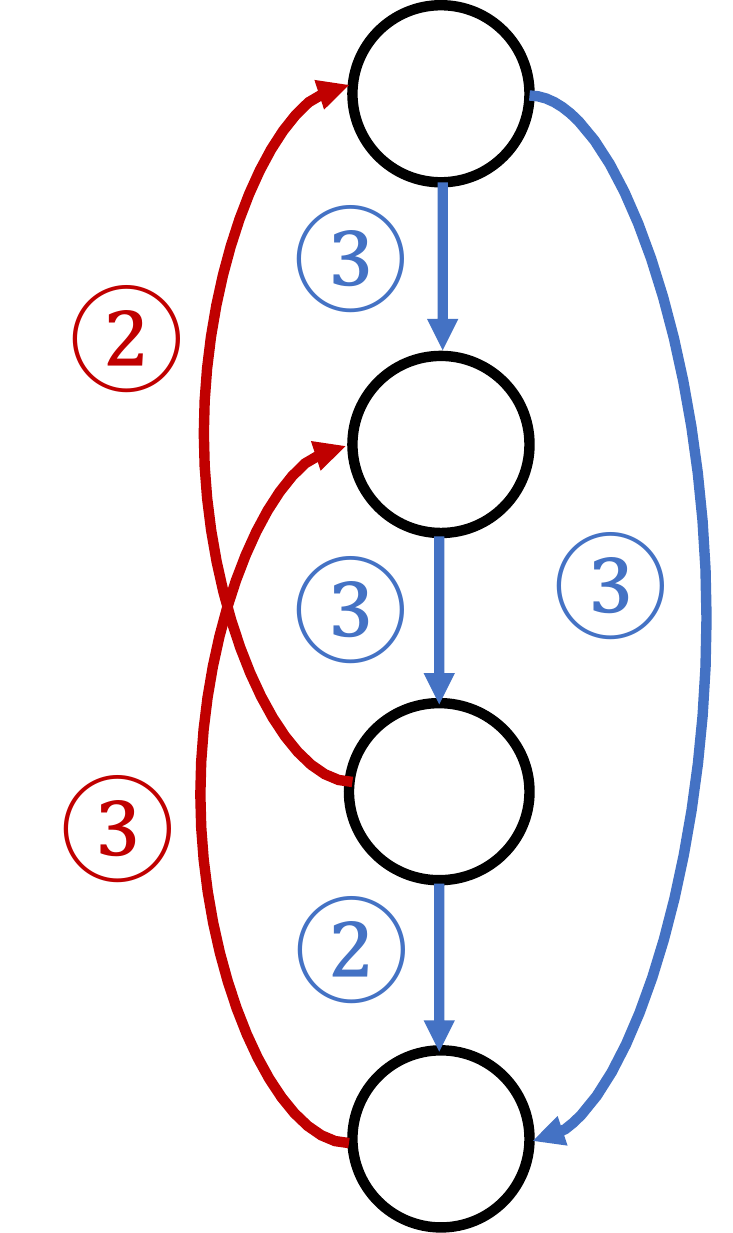}
\\
{ Acc: 93.11$\%$} & 
{ Acc: 70.20$\%$}&
{ Acc: 52.58$\%$}&
{ Acc: 93.82$\%$} &
{ Acc: 73.31$\%$ }& 
{Acc: 55.41$\%$}
\\
{ (a) C10-Fw} & 
{(b)  C100-Fw }&
{ (c)  Tiny-Fw }&
{ (d) C10-Bw} &
{(e) C100-Bw  }& 
{(f) Tiny-Bw }\\
\end{tabular}
\caption{Searched architecture examples (forward and backward configuration) for three benchmarks. \textcolor{blue}{Blue} and \textcolor{red}{red} arrows denote forward connection and backward connection, respectively. The number on each arrow represents operations introduced in Fig. \ref{fig:method:cellbased-nas}.
}
\label{fig:exp:best_architecturecell}
\end{center}
\end{figure}

\subsection{Performance Comparison}
\vspace{-2mm}
Table~\ref{table:performance} shows the performance comparison between our SNASNet founded by the proposed NAS algorithm and previous SNN models on three benchmarks.
As our NAS approach has randomness, we run the same configuration 5 times and report the mean and standard deviation.
In the table, ``SNASNet-Fw" refers to our searched model with only forward connections 
and ``SNASNet-Bw" denotes our searched model with both forward and backward connections.
SNASNet-Fw achieves comparable performance with the previous works with extremely small timesteps.
For example,  our searched model achieves $70.06\%$ with timestep 5 on CIFAR100, which is similar to the VGG16 model performance from Rathi \etal \cite{rathi2021diet}.
Note that, for CIFAR10, which is a relatively simple dataset, a few methods yield marginally better performance than SNASNet-Fw. 
Interestingly, compared to SNASNet-Fw, SNASNet-Bw improves the performance by 0.61$\%$, 2.98$\%$, and 1.79$\%$, for CIFAR10, CIFAR100, and TinyImageNet, respectively. We note that SNASNet-Bw yields SOTA results across all datasets with only 5/8 timesteps.
The results support our assertion that the representation power of SNNs can be enhanced by passing information through backward connections where temporal information is further exploited.
We also illustrate the example of searched architecture cell found by our proposed NAS algorithm for each dataset in Fig. \ref{fig:exp:best_architecturecell}.
Recently, Shu \etal \cite{shu2019understanding} show that fast convergence ANN architectures bring  smooth loss landscape and accurate
gradient information, resulting in high test accuracy.
We also found that our searched SNN architectures achieve fast convergence with high test accuracy, as shown in Fig. \ref{fig:method:training_convergence}.
By using this early stage information, there is a possibility of applying an evolutionary algorithm \cite{real2019regularized} to SNN searching in future works.

\begin{figure}[t]
\begin{center}
\def\arraystretch{0.5}
\begin{tabular}{@{}c@{}c@{\hskip 0.03\linewidth}c@{}c@{}c@{}c@{}c}
\includegraphics[width=0.145\linewidth]{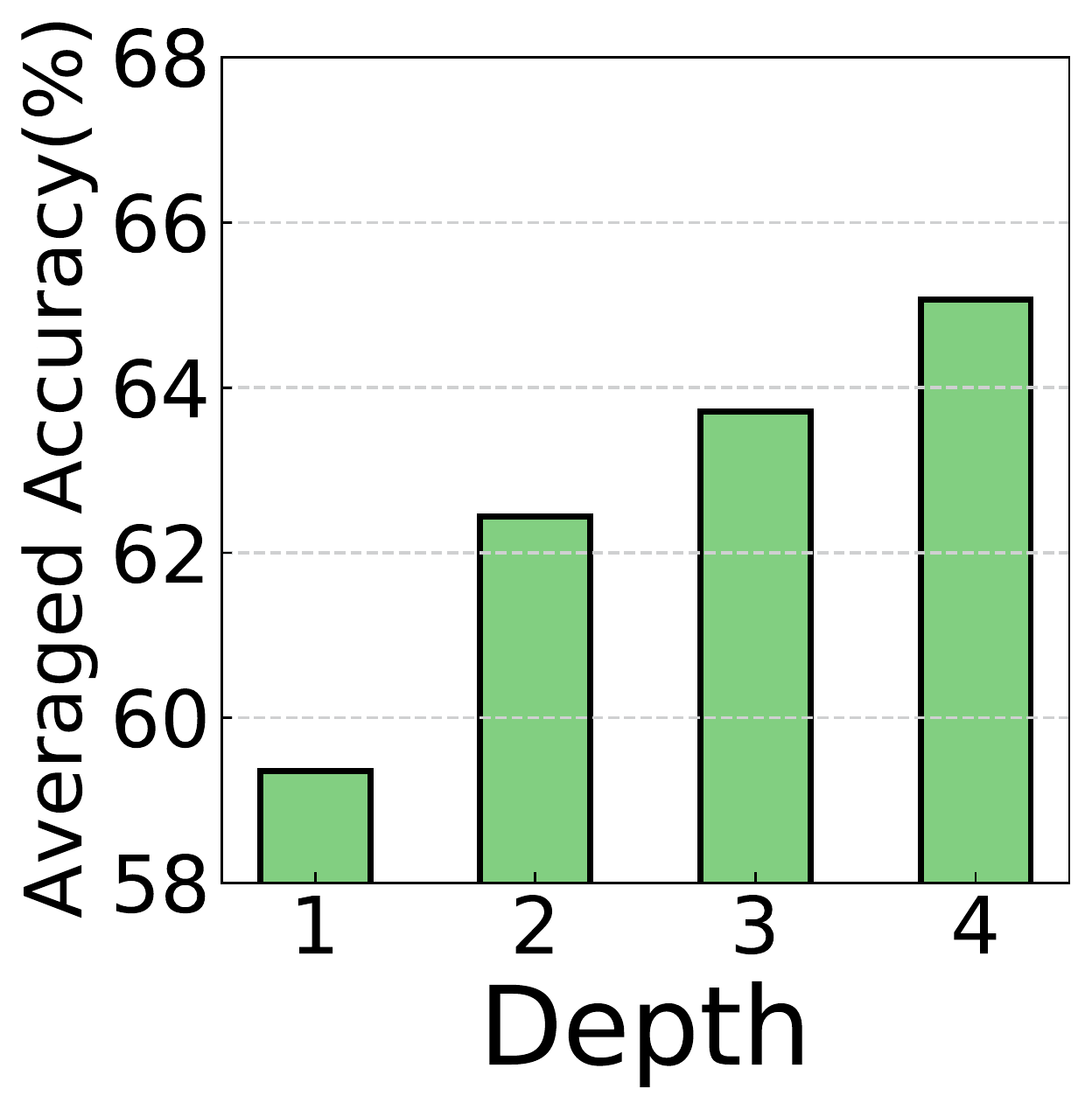}&
\includegraphics[width=0.13\linewidth]{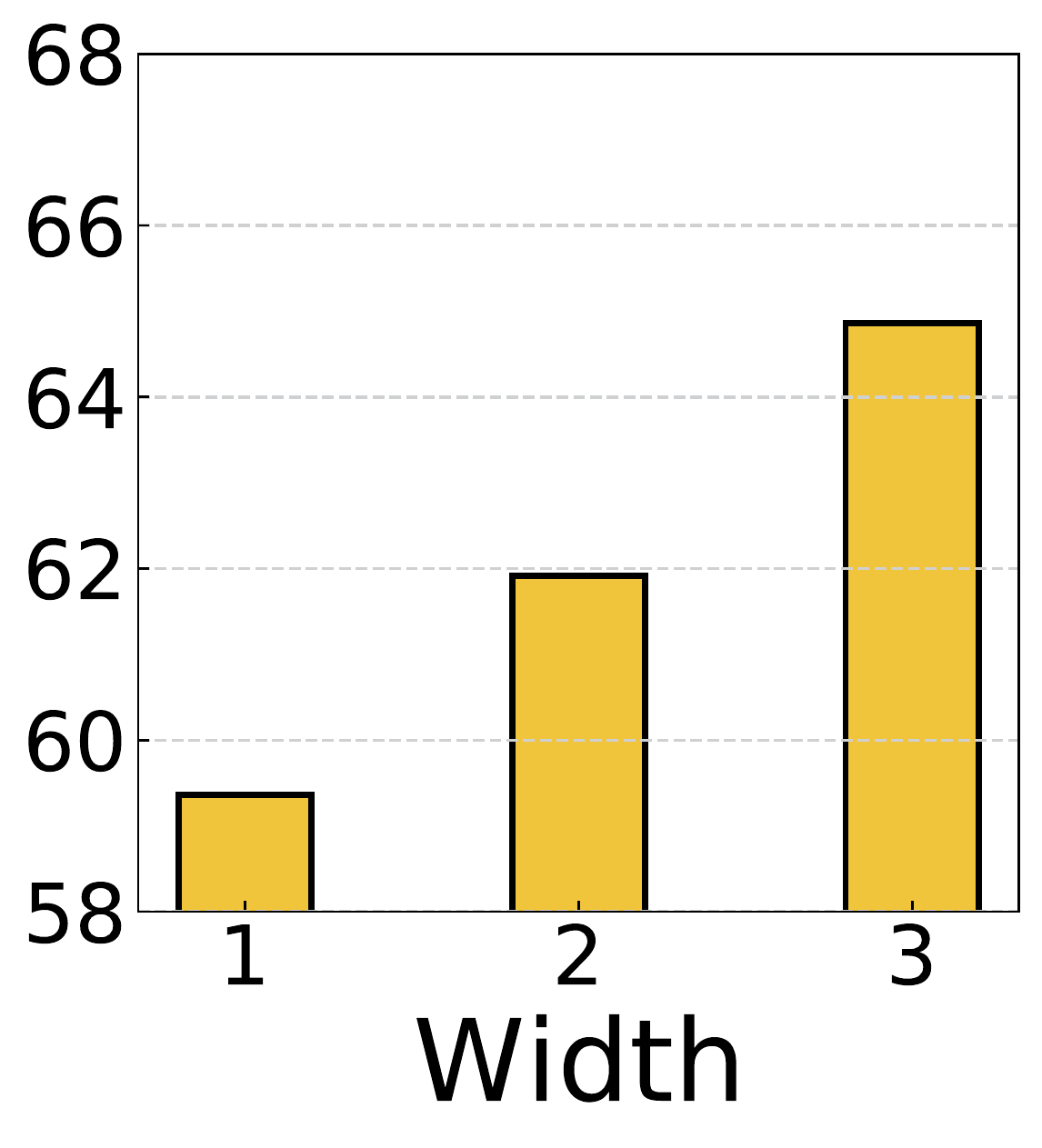}&
\includegraphics[width=0.145\linewidth]{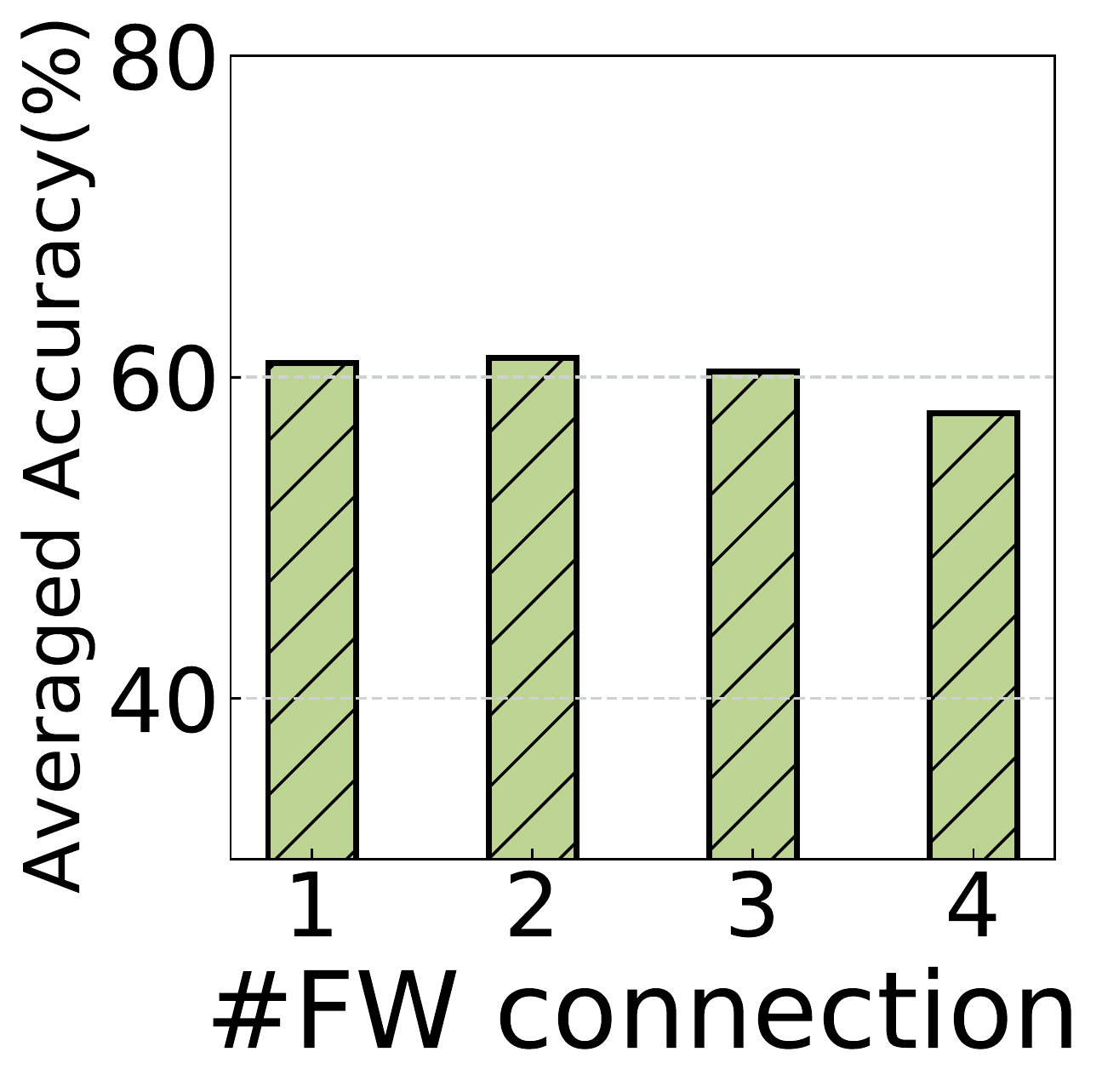} &
\includegraphics[width=0.13\linewidth]{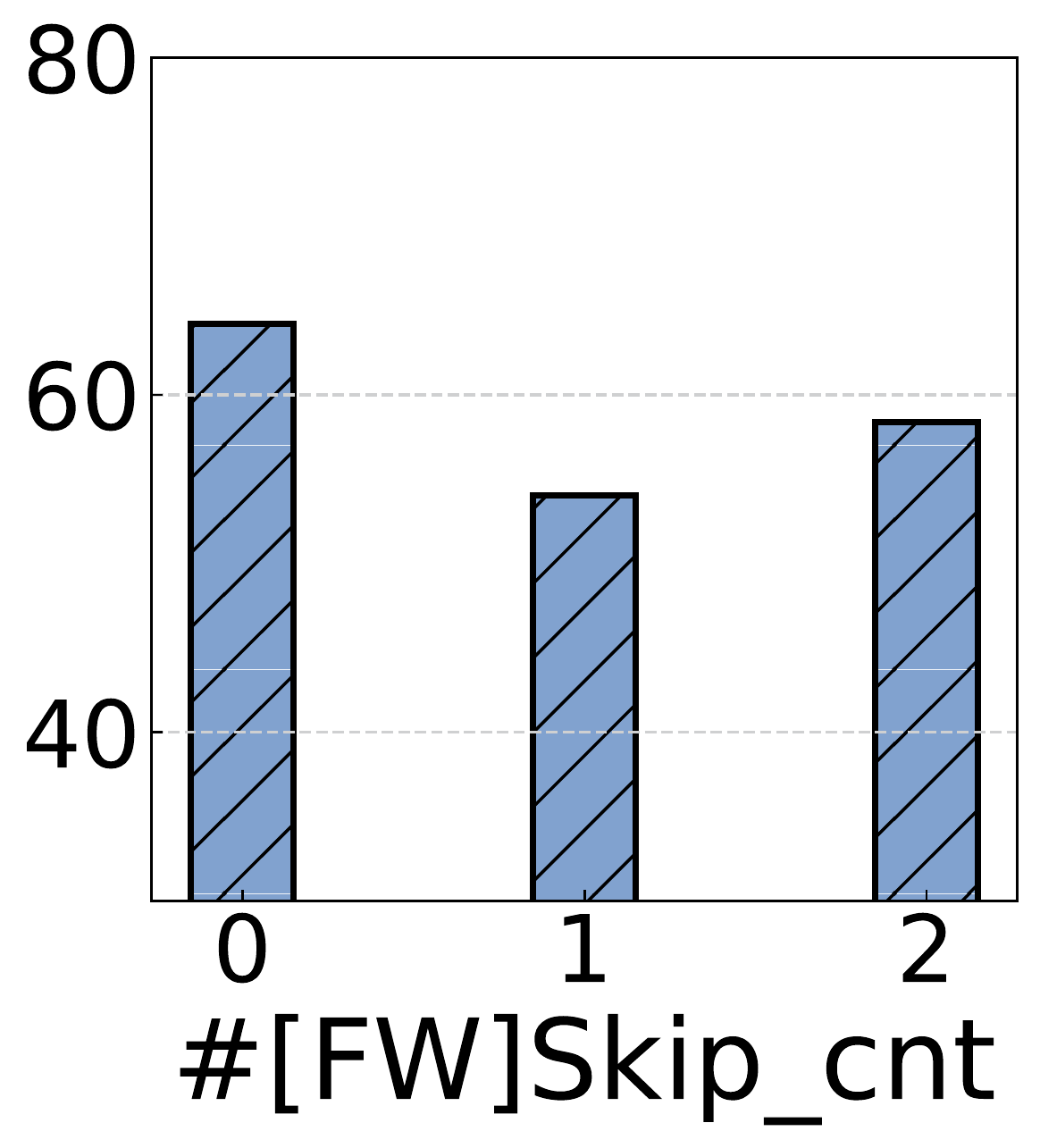}&
\includegraphics[width=0.13\linewidth]{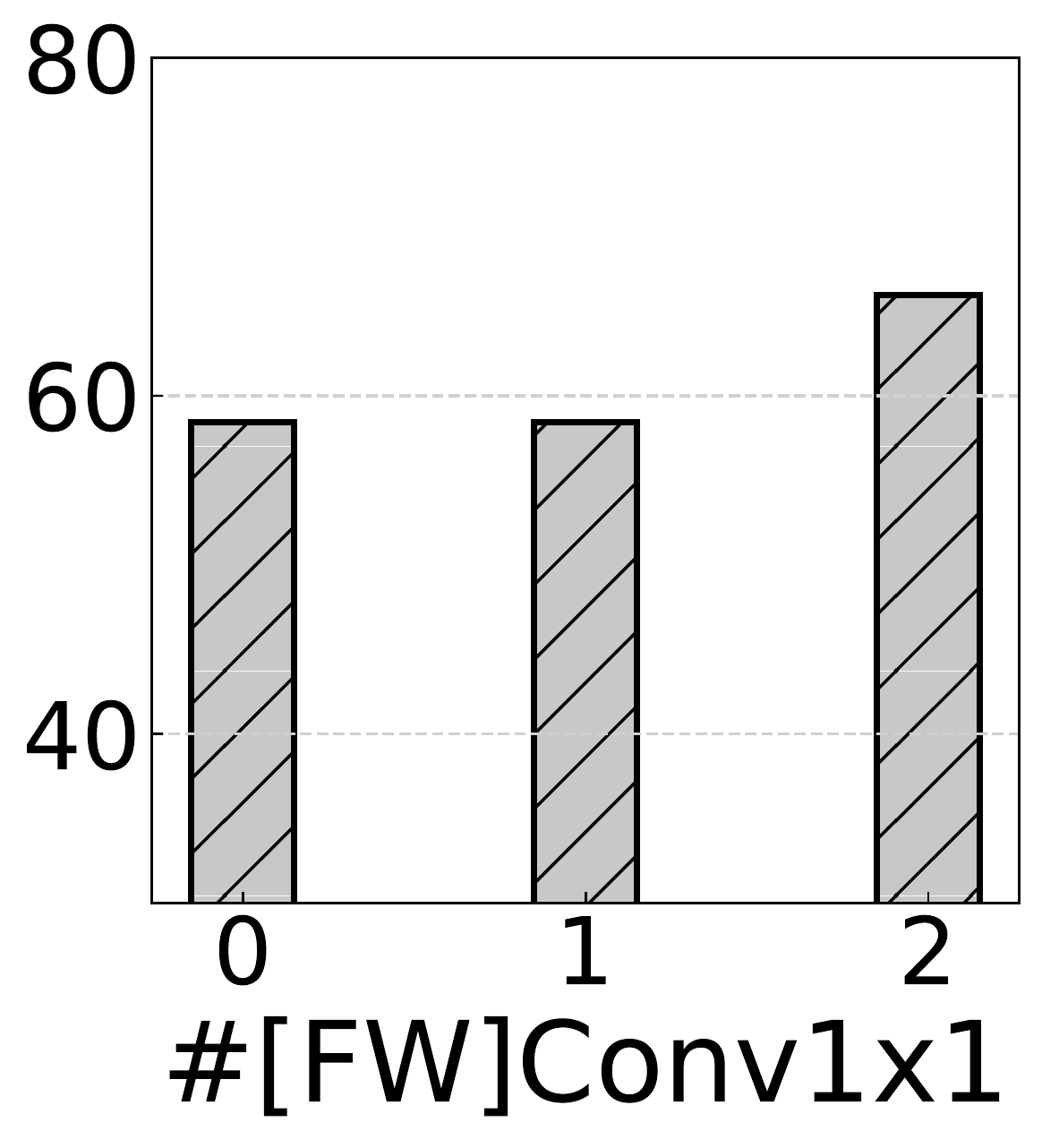}&
\includegraphics[width=0.13\linewidth]{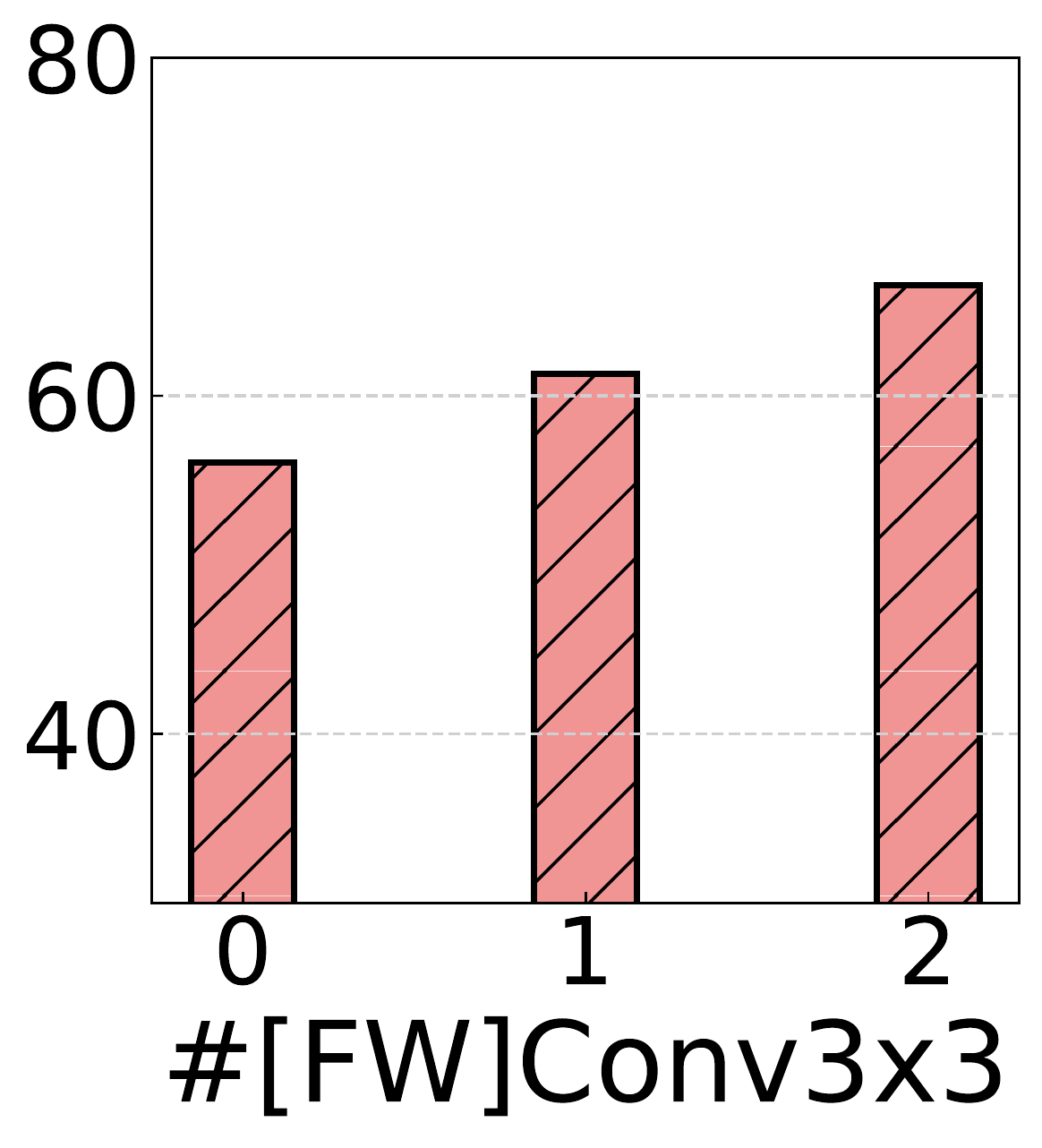}&
\includegraphics[width=0.13\linewidth]{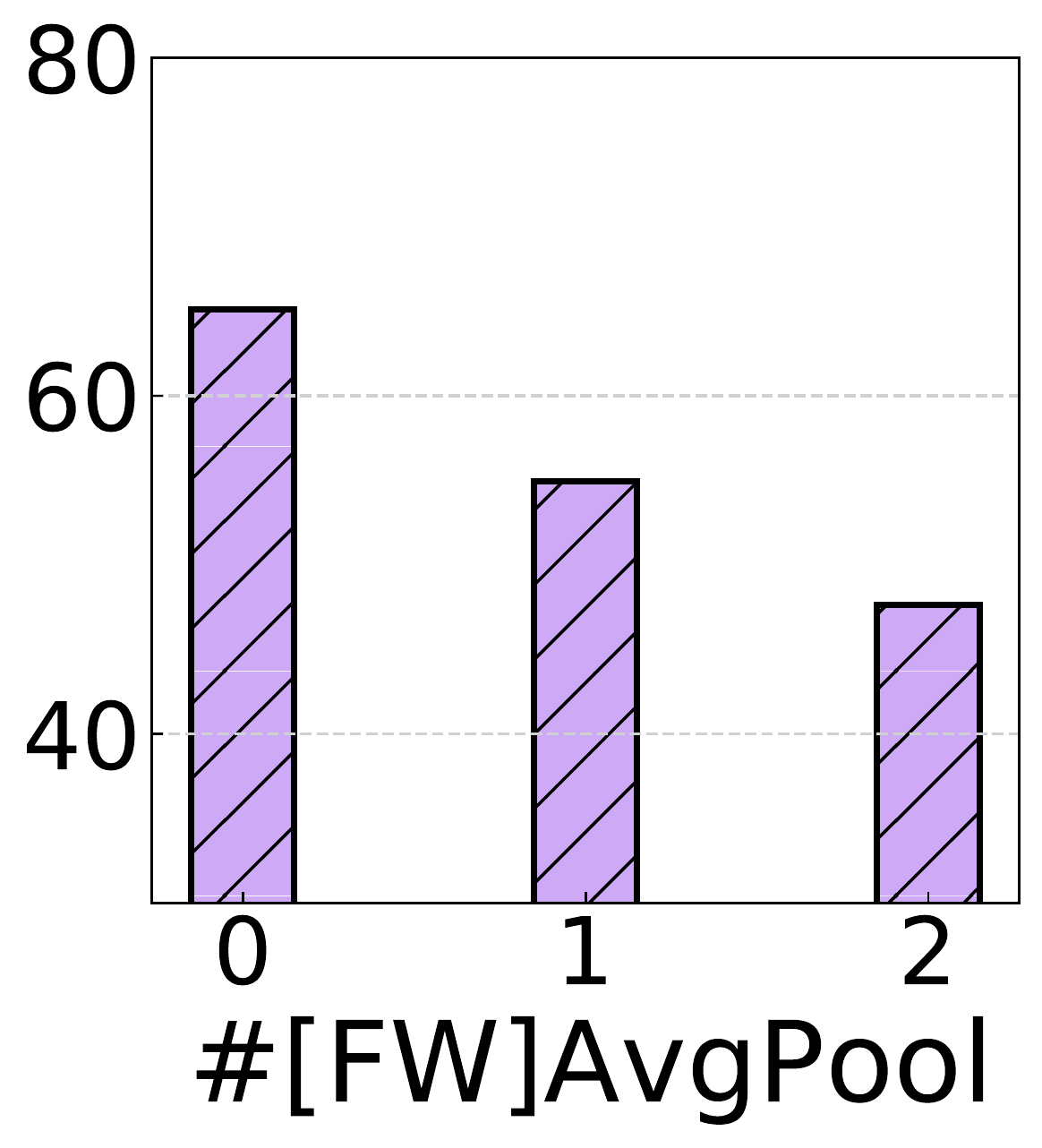}
\\
\includegraphics[width=0.14\linewidth]{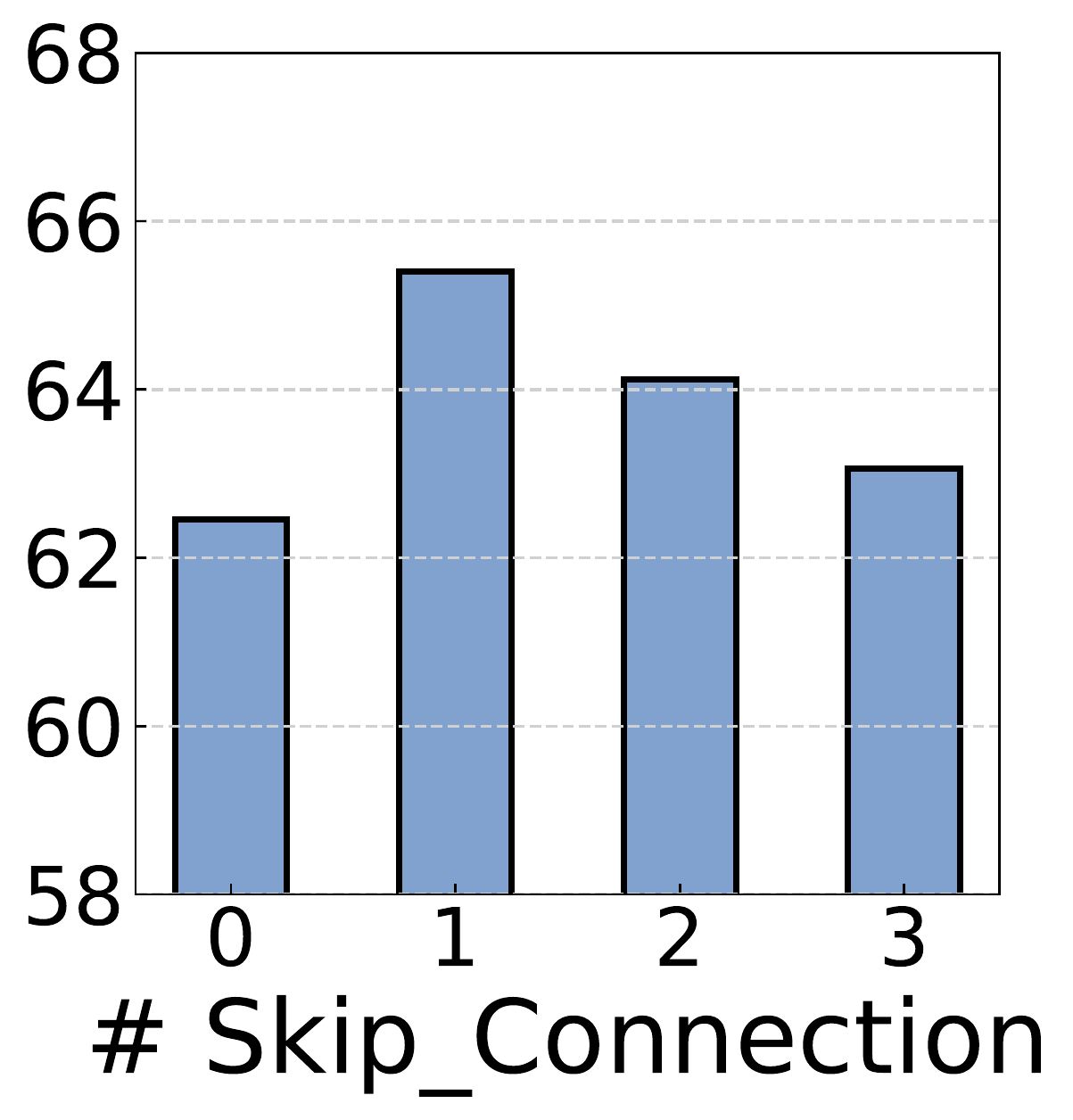}&
\includegraphics[width=0.13\linewidth]{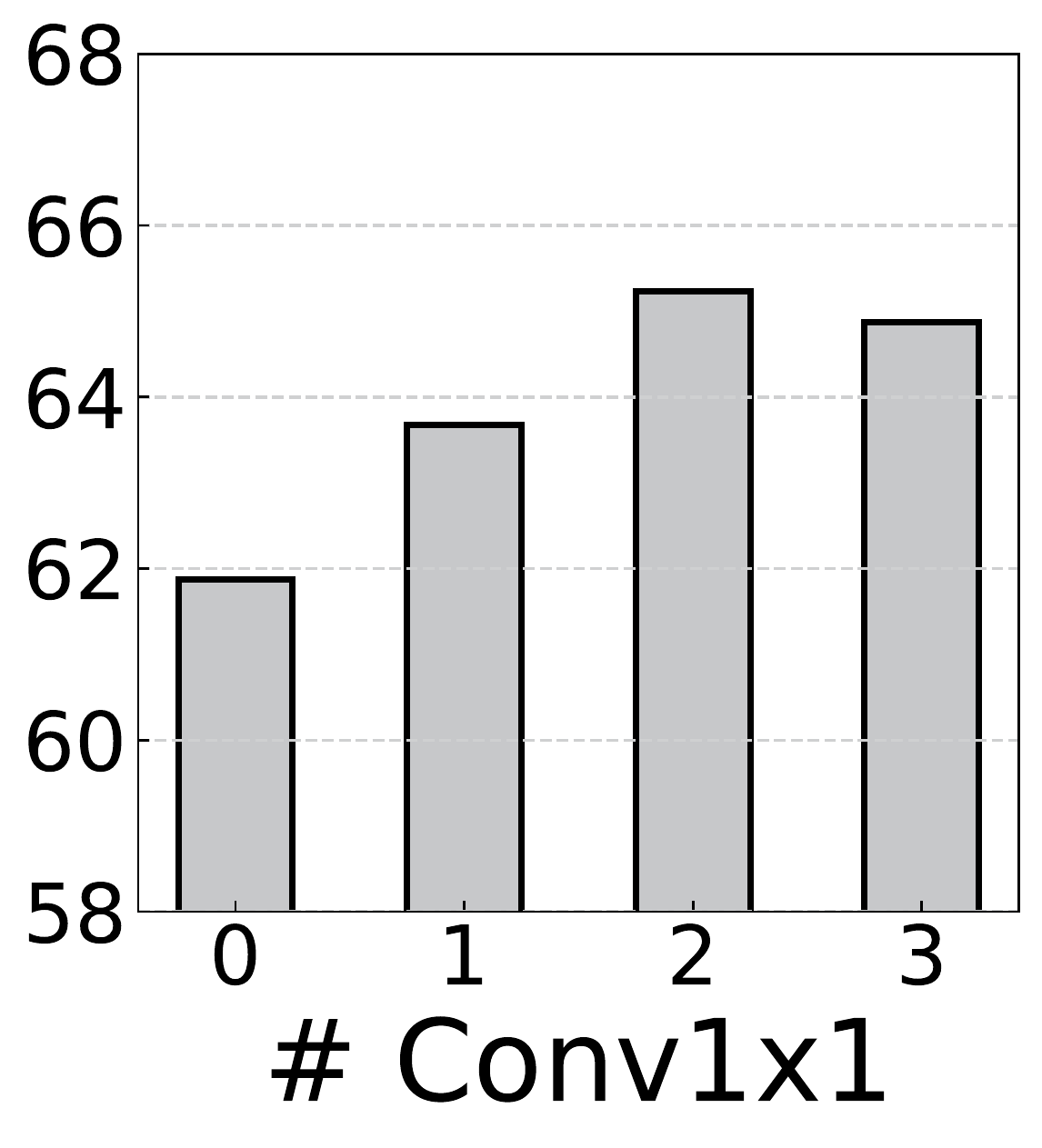}&
\includegraphics[width=0.145\linewidth]{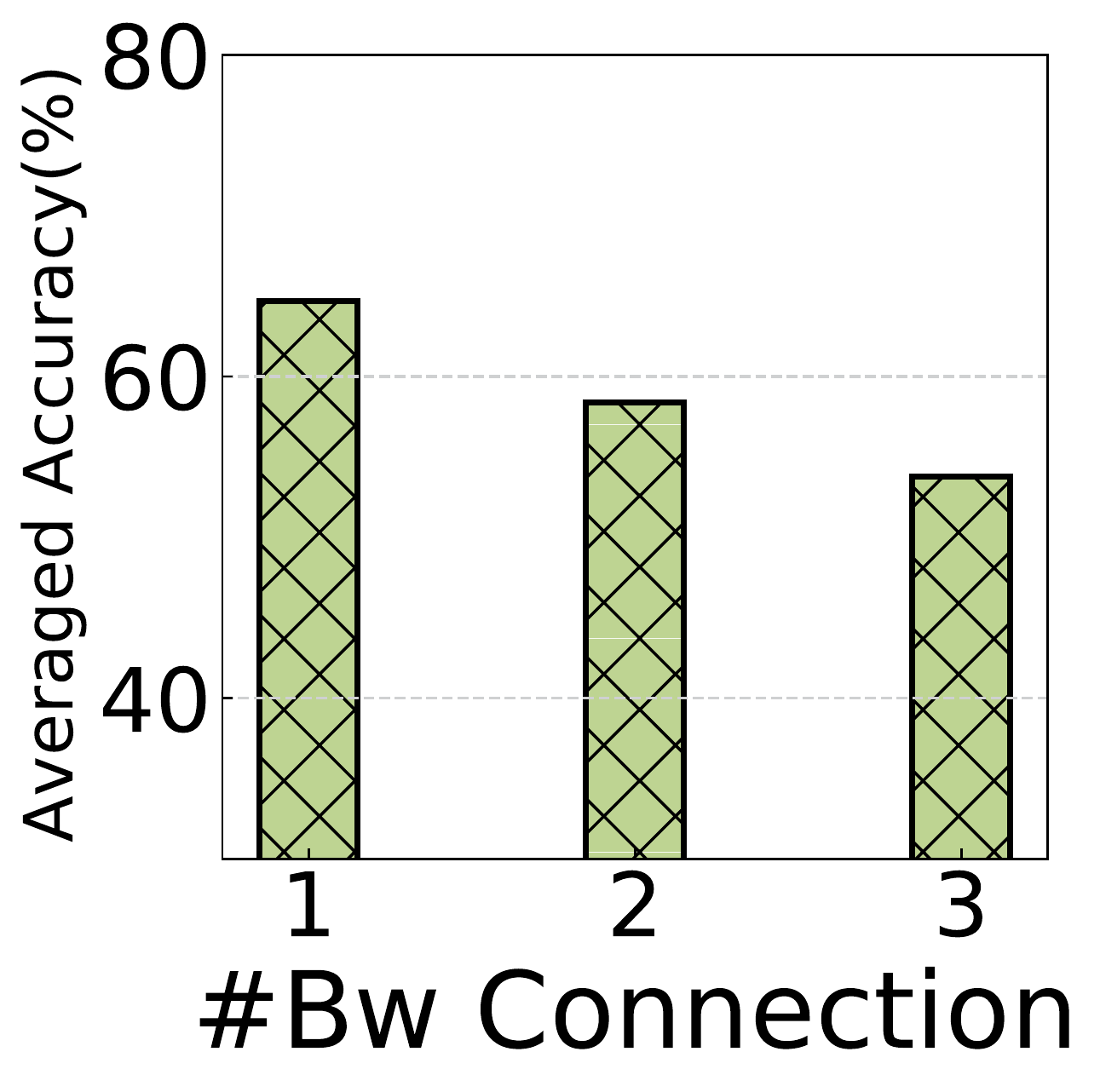} &
\includegraphics[width=0.13\linewidth]{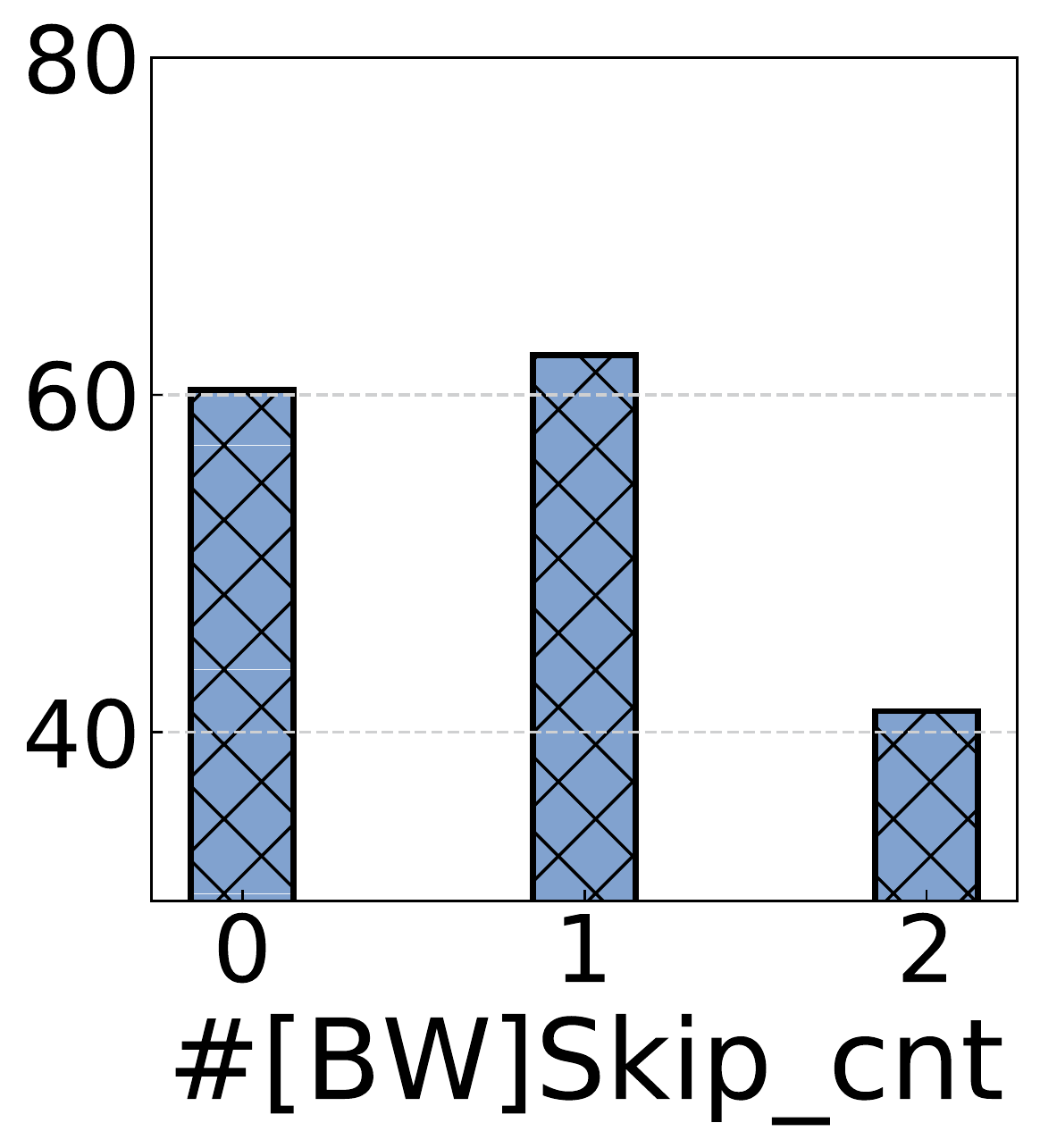}&
\includegraphics[width=0.13\linewidth]{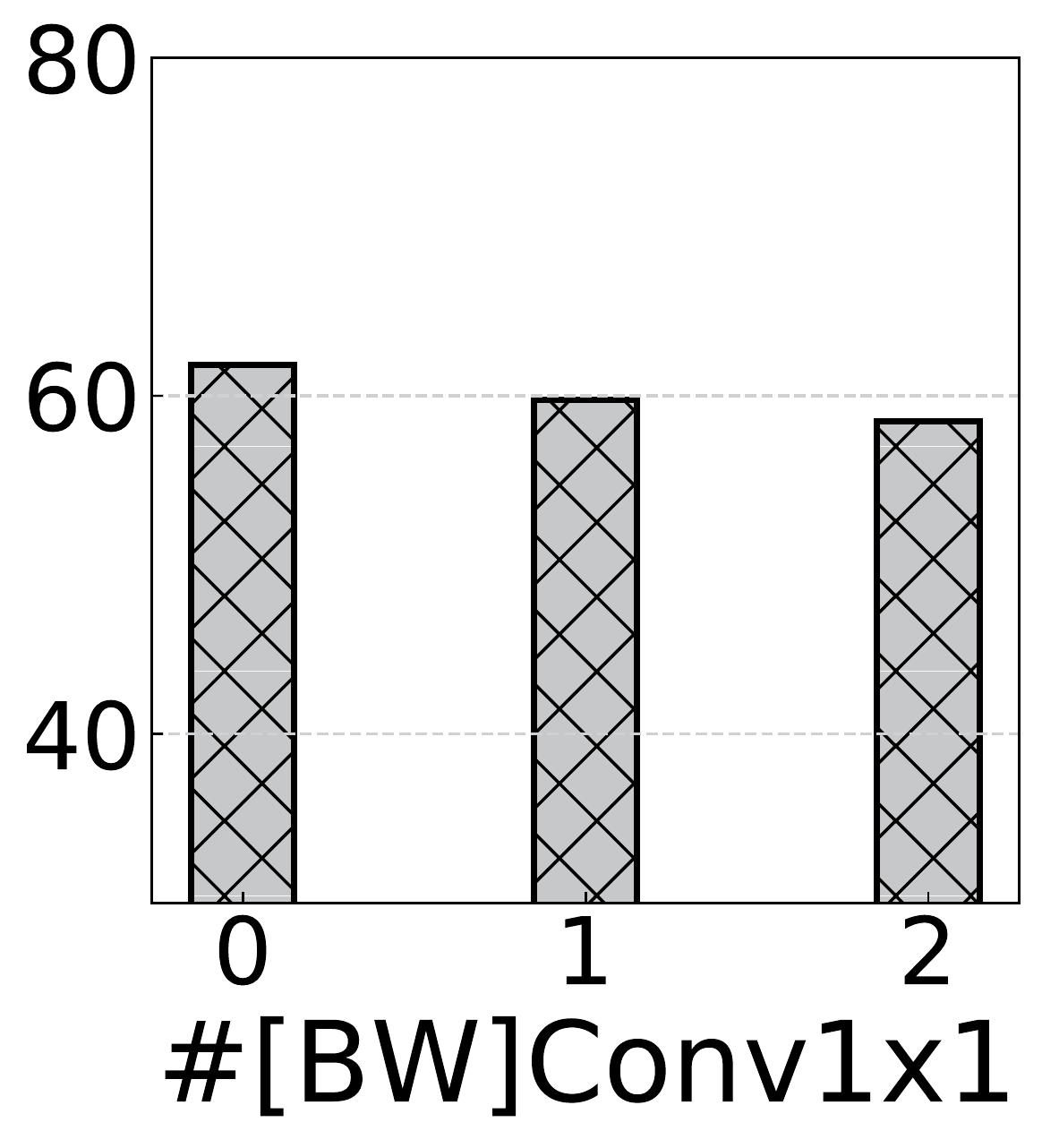}&
\includegraphics[width=0.13\linewidth]{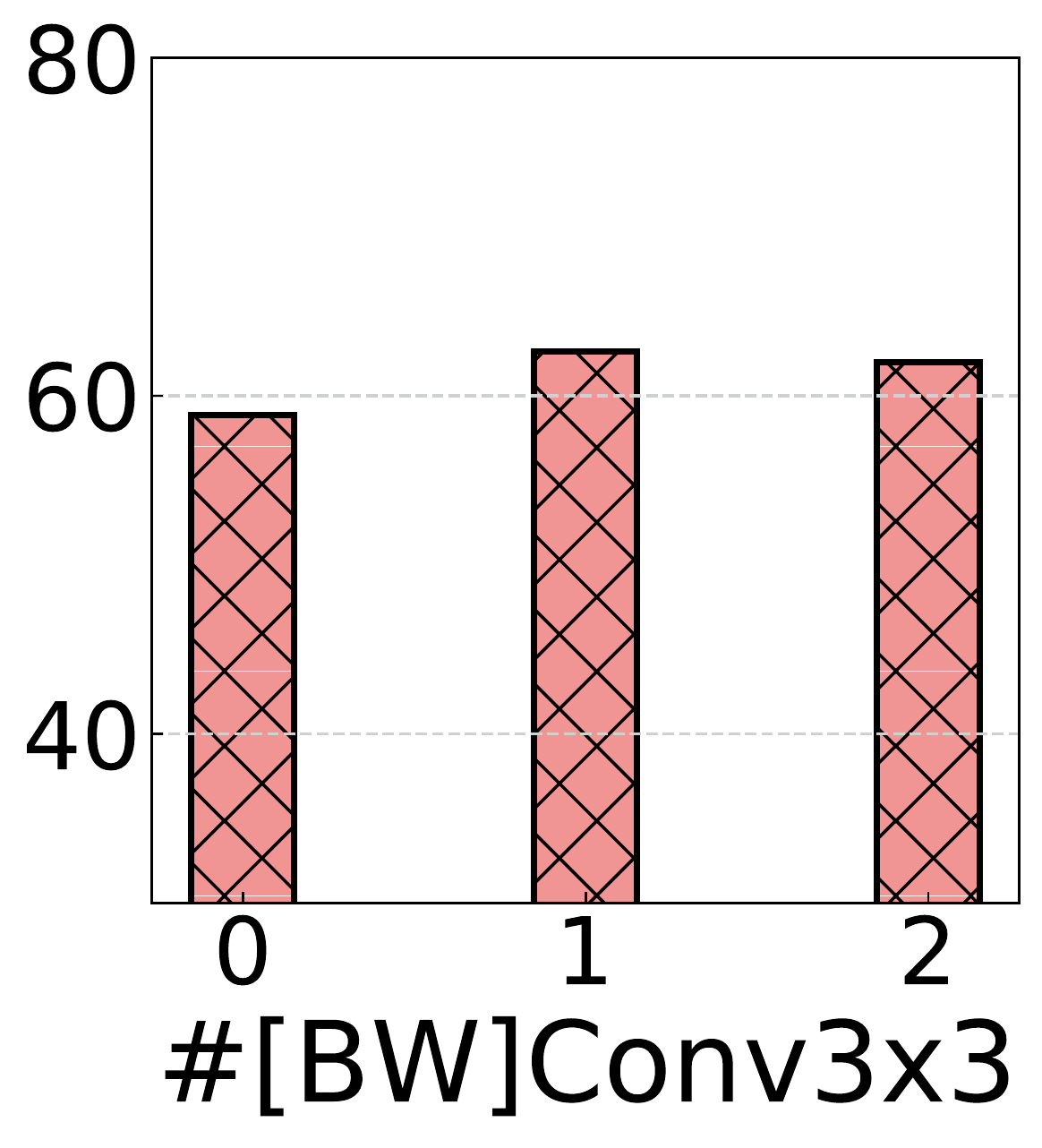}&
\includegraphics[width=0.13\linewidth]{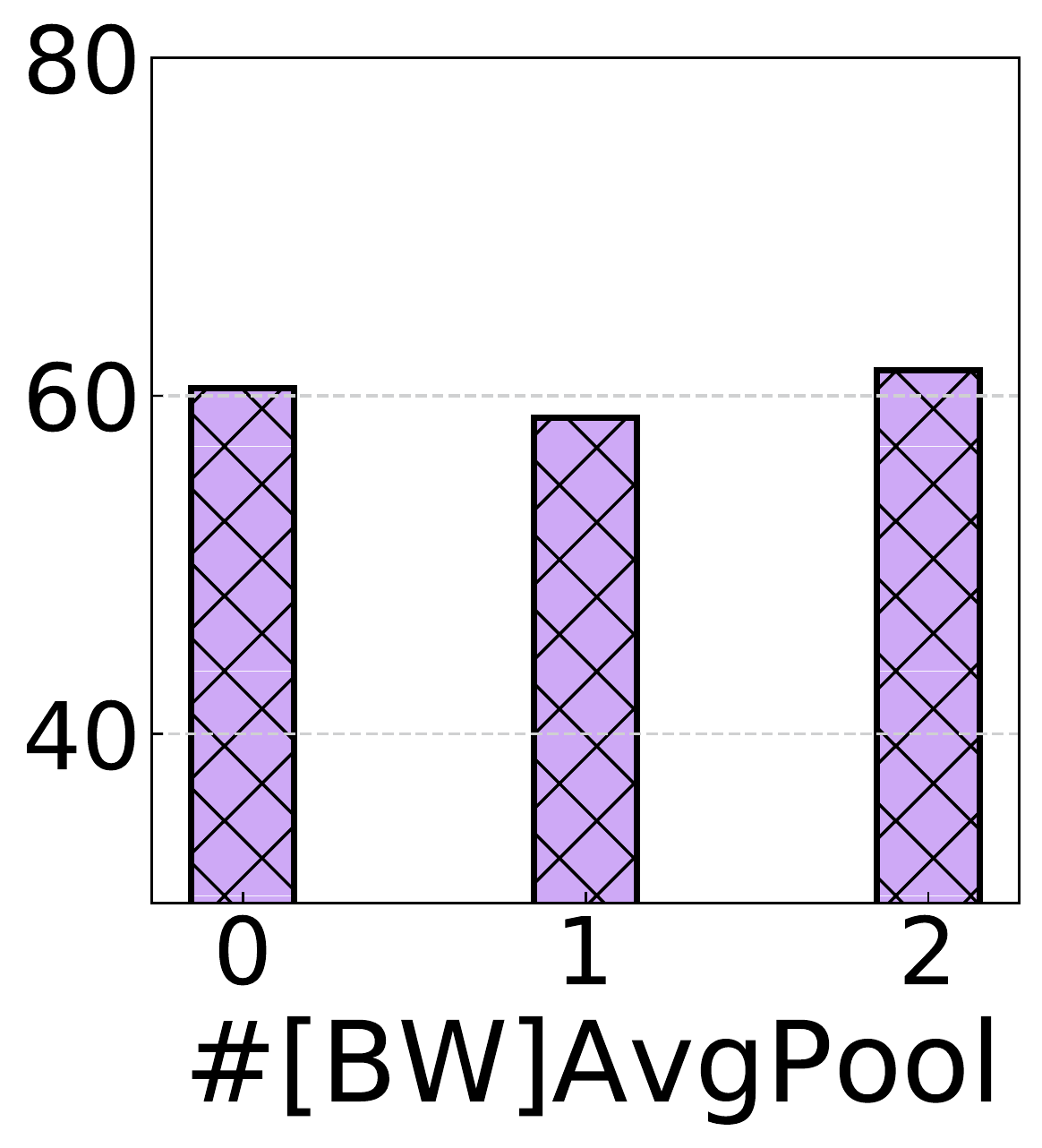}
\\
\includegraphics[width=0.133\linewidth]{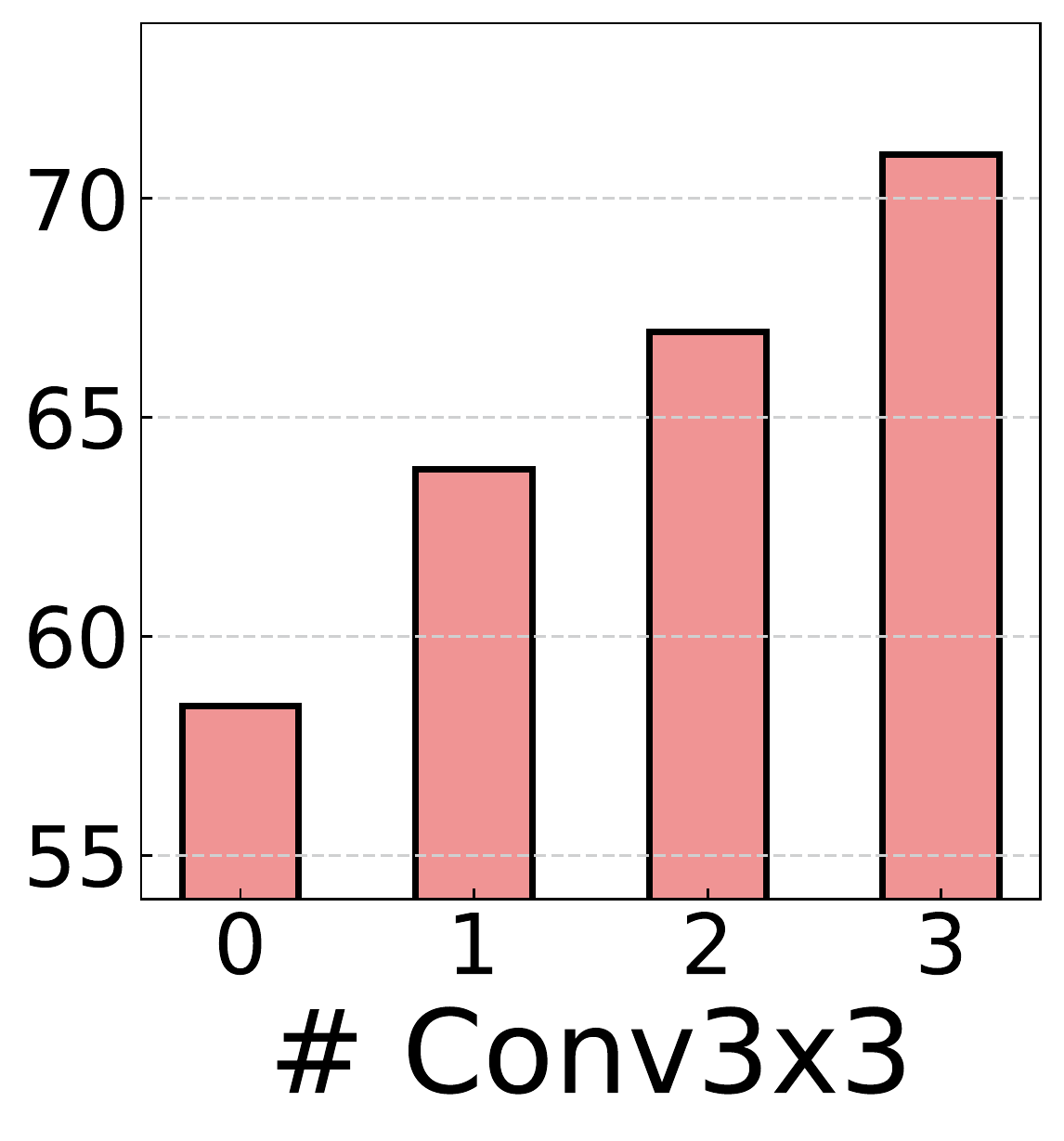}&
\includegraphics[width=0.13\linewidth]{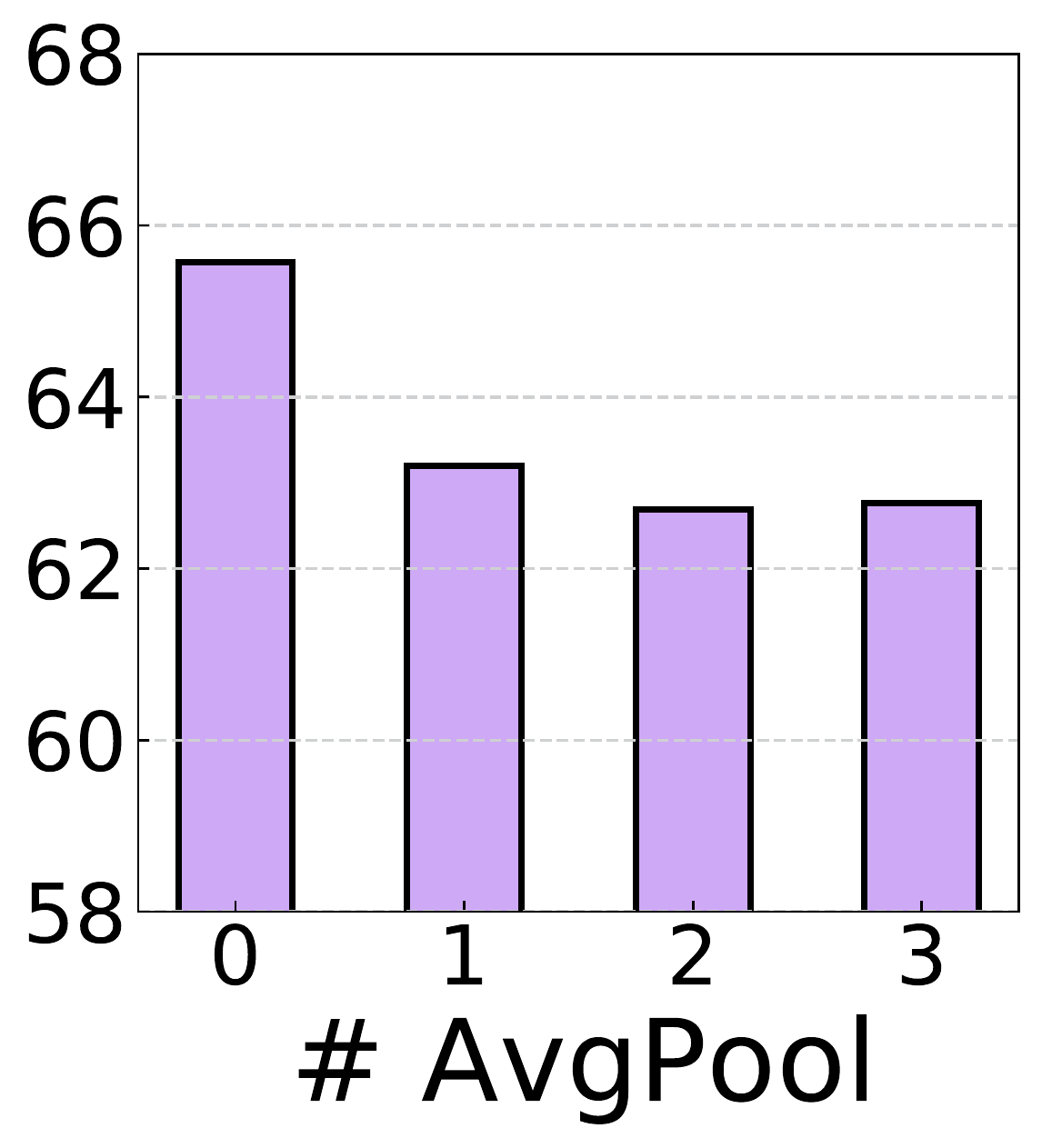}&
\includegraphics[width=0.135\linewidth]{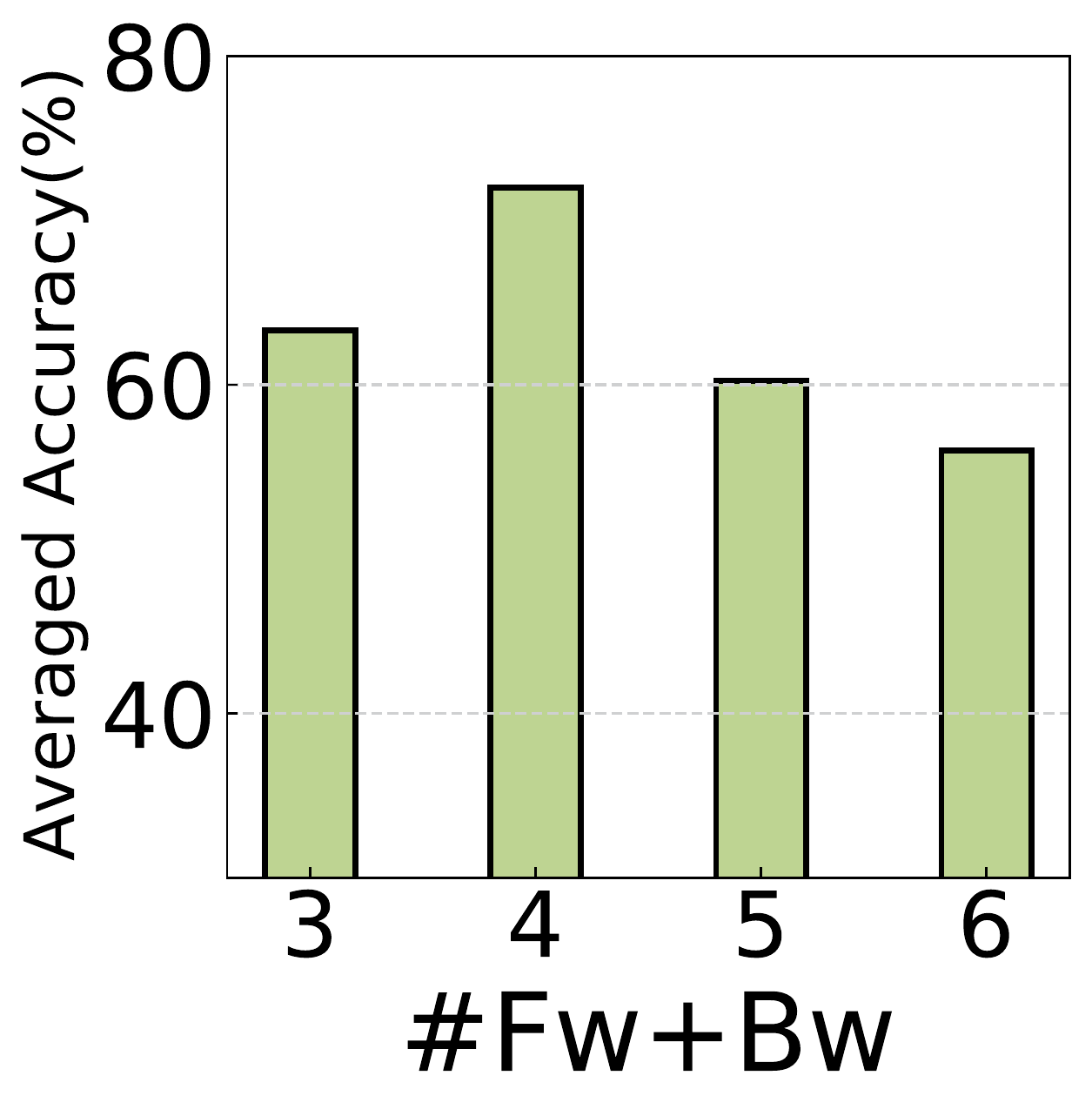}& 
\includegraphics[width=0.13\linewidth]{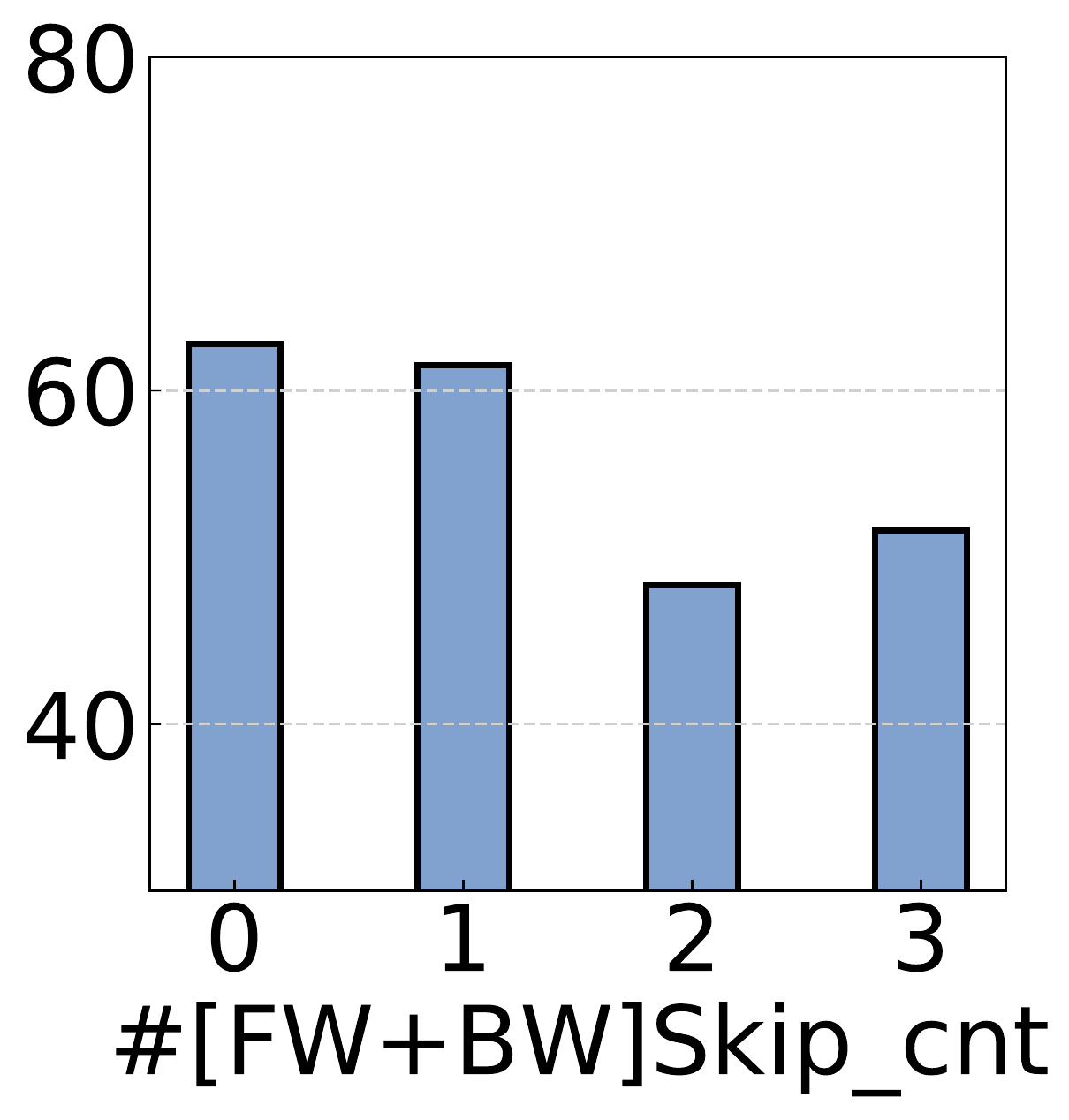}&
\includegraphics[width=0.13\linewidth]{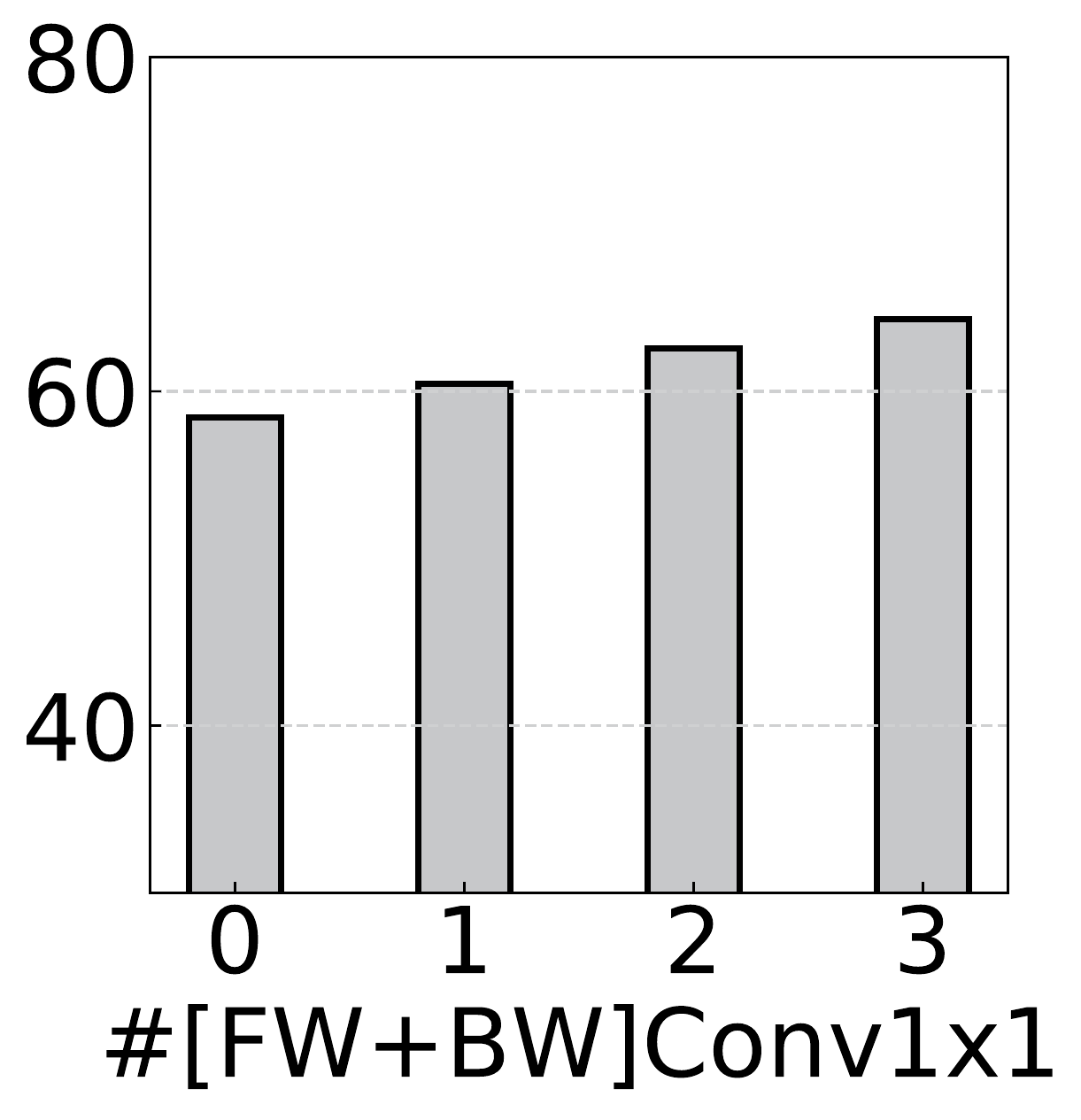}&
\includegraphics[width=0.13\linewidth]{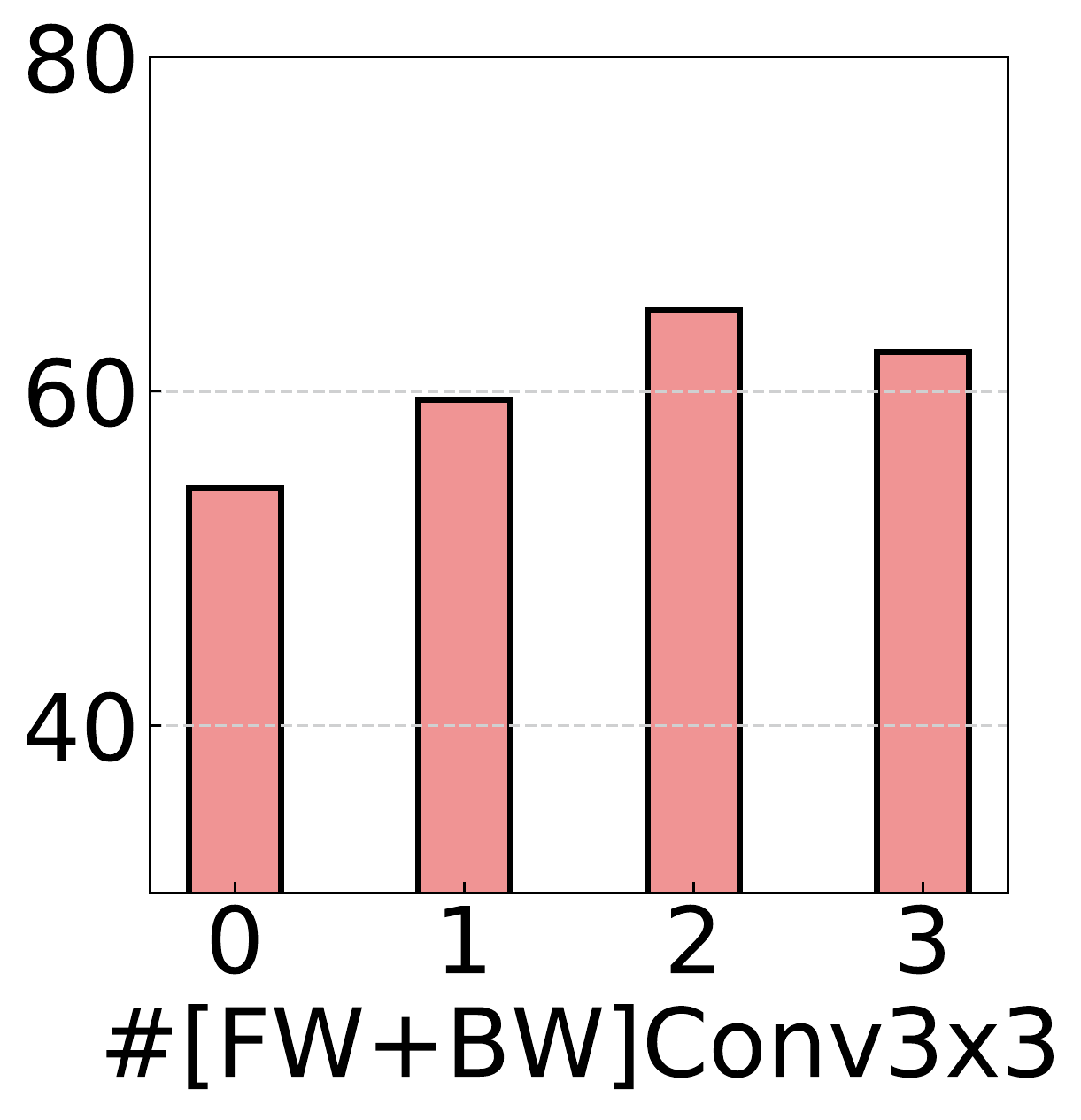}&
\includegraphics[width=0.13\linewidth]{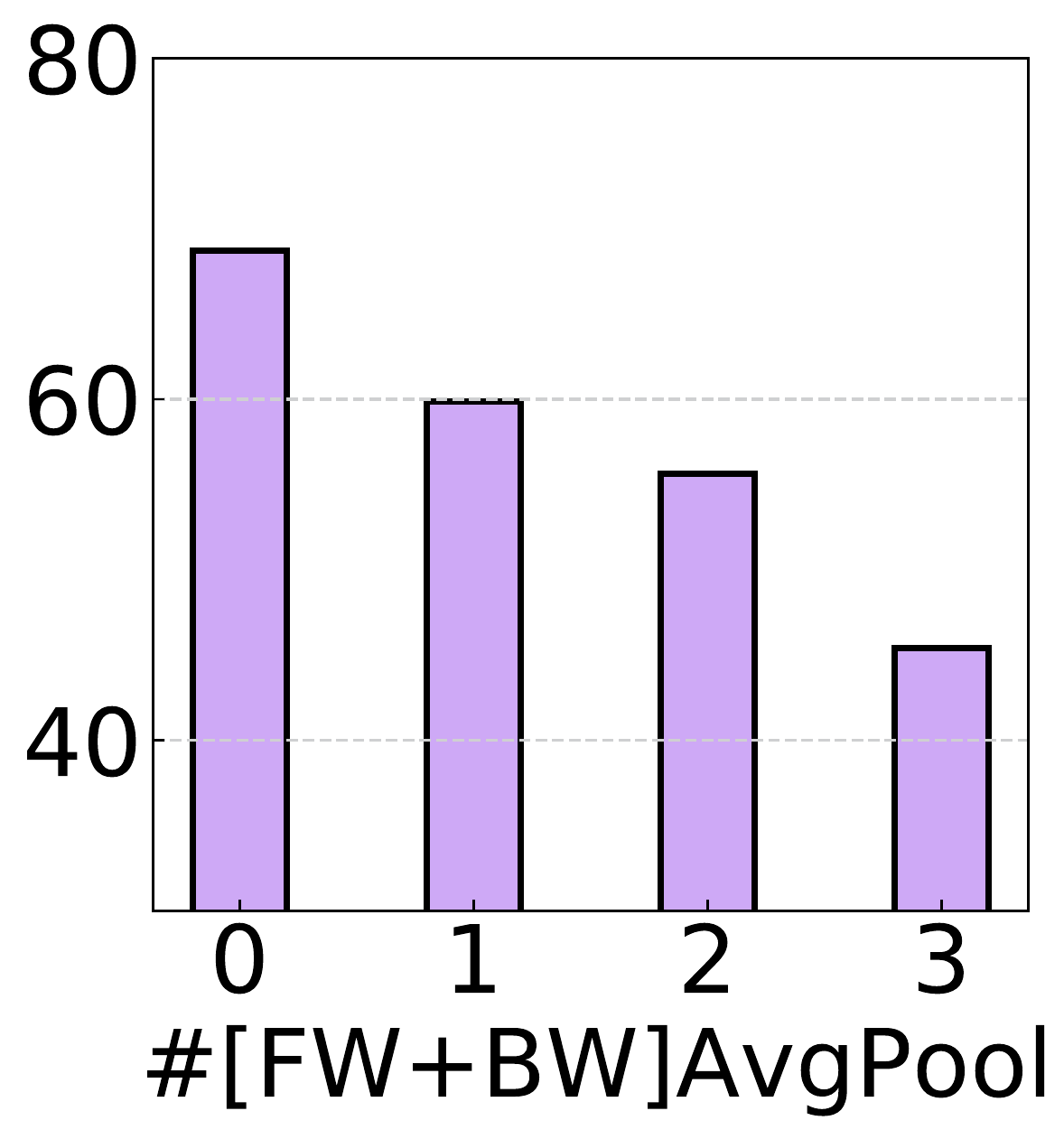} 
\\
\multicolumn{2}{c}{Forward connections} & \multicolumn{5}{c}{Forward + Backward connections} 
\end{tabular}
\caption{
Accuracy vs. cell attributes. \textbf{1st $\&$ 2nd columns:} Searching Fw connection only. 
\textbf{3rd$\sim$7th columns:}  Searching both Fw and Bw connections. 1st, 2nd, 3rd rows show statistics of Fw, Bw,  Fw+Bw, respectively. 
}
\label{fig:observation}
\end{center}
\end{figure}

\subsection{Experimental Analysis}

\noindent\textbf{Observations from Searched Cells. }
We provide several observations obtained from our searching algorithm.
To this end, in Fig. \ref{fig:observation}, we ran 100 random searches on CIFAR100 
and provide averaged accuracy with respect to the number of forward connections (Fw), backward connections  (Bw), skip connections, Conv $3\times3$, Conv $1\times1$, Average pooling.
The key observations are as follows.
For \textbf{SNASNet with only {forward connections}}, 
{(1)} a deeper and wider cell improves performance, which implies that scaling up SNN is important (1st $\&$ 2nd columns in Fig. \ref{fig:observation}).
{(2)} Convolutional layers are important for getting higher performance. On the other hand, average pooling is not preferred for SNNs.
For \textbf{SNASNet with both {forward and backward connections} (3rd$\sim$7th columns in Fig. \ref{fig:observation})},
{(1)} The trend of forward connections (1st row) also prefer convolutional layers, which is similar to that of 1st row in Fig. \ref{fig:observation}.
{(2)} As shown in the 2nd row, a small number of backward connections are preferred.
Also, the type of connections does not affect the accuracy except for skip connection.
{(3)} More than 2 backward skip connections degrade accuracy significantly. This implies that feedback without transformation (\eg,  convolutional or pooling operation) deteriorates representation of SNNs.
{(4)} Considering both forward and backward connections (3rd row), the total number of connections should be carefully designed. 

\begin{table}[t]
\addtolength{\tabcolsep}{0.5pt}
\centering
\caption{Transferability study of founded architectures.  $\Delta$Acc denotes the performance change compared to the best performed architecture shown in Fig. \ref{fig:exp:best_architecturecell}.}
\resizebox{0.80\textwidth}{!}{%
\begin{tabular}{ccccc}
\toprule
\:\:\:\:Celltype\:\:\:\: &\:\:\:\: Searching dataset\:\:\: \: & \:\:\:\:Train/test dataset \:\:\:\: & \:\:\:\:Accuracy ($\%$) \:\:\:\:&\:\: \:\:$\Delta$Acc ($\%$)\:\:\:\:\\
\midrule
Forward & CIFAR10 &  CIFAR100 &  69.98 & -0.22   \\
Forward & CIFAR10 &  TinyImageNet & 52.02   & -0.56      \\
Forward & CIFAR100 &  CIFAR10 & 93.12      & +0.01   \\
Forward & CIFAR100 &  TinyImageNet & 52.28    & -0.30       \\
Forward & TinyImageNet &  CIFAR10 & 93.16   & +0.05      \\
Forward & TinyImageNet &  CIFAR100 & 70.31   & +0.11       \\
\midrule
Backward & CIFAR10 &  CIFAR100 & 73.19     & -0.12    \\
Backward & CIFAR10 &  TinyImageNet & 54.61  & -0.80        \\
Backward & CIFAR100 &  CIFAR10 & 93.73   & -0.09      \\
Backward & CIFAR100 &  TinyImageNet & 56.00  & +0.59      \\
Backward & TinyImageNet &  CIFAR10 & 93.64    & -0.18     \\
Backward & TinyImageNet &  CIFAR100 & 73.14   & -0.17       \\
\bottomrule
\end{tabular}%
}
\label{table:exp:transferability}
\end{table}

\begin{figure}[t]
\noindent\begin{minipage}{\textwidth}
\begin{minipage}{0.48\textwidth}
\centering
\begin{figure}[H]
\begin{center}
\def\arraystretch{0.5}
\begin{tabular}{@{}c@{\hskip 0.01\linewidth}c@{}c}
\includegraphics[width=0.49\linewidth]{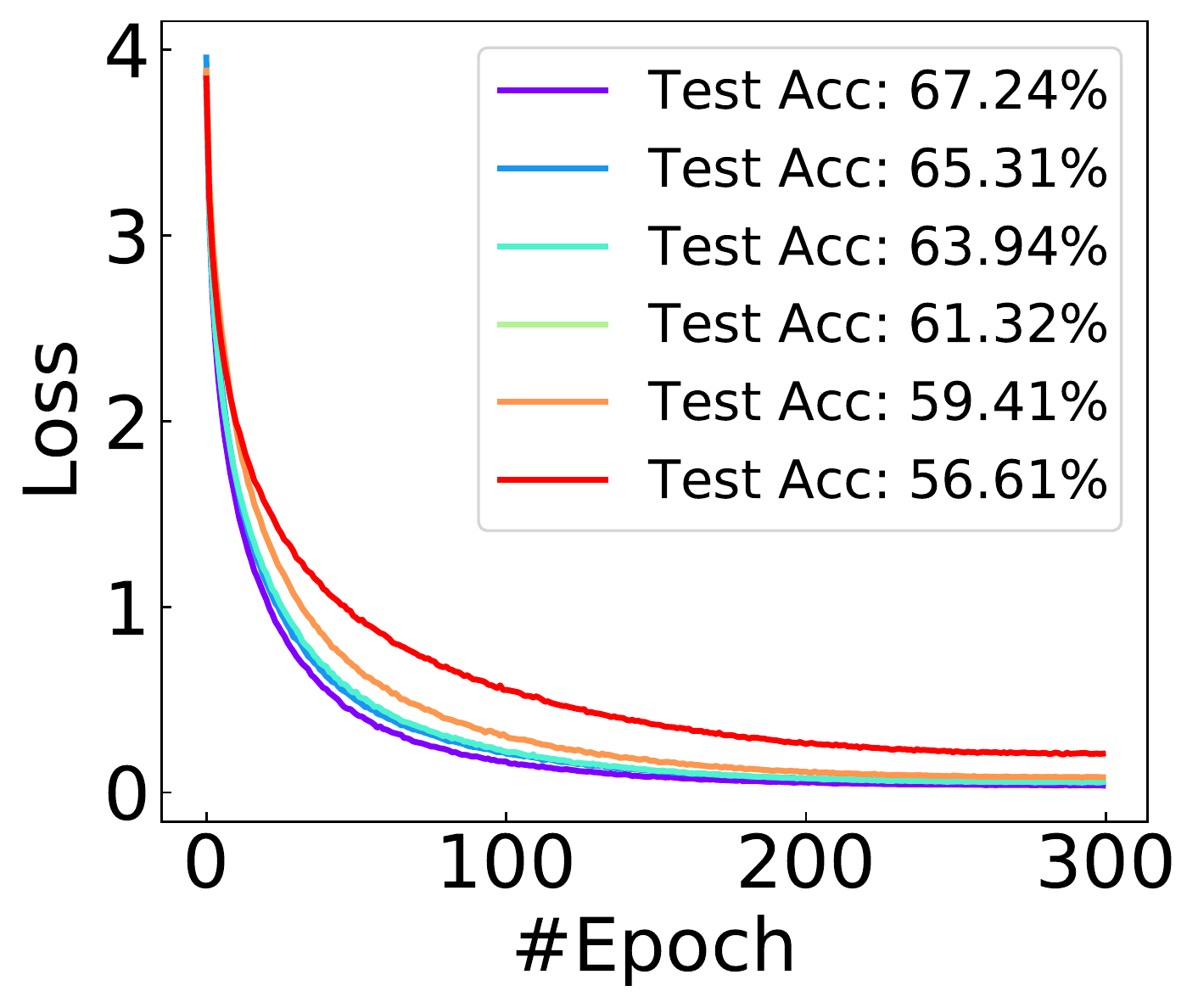} &
\includegraphics[width=0.52\linewidth]{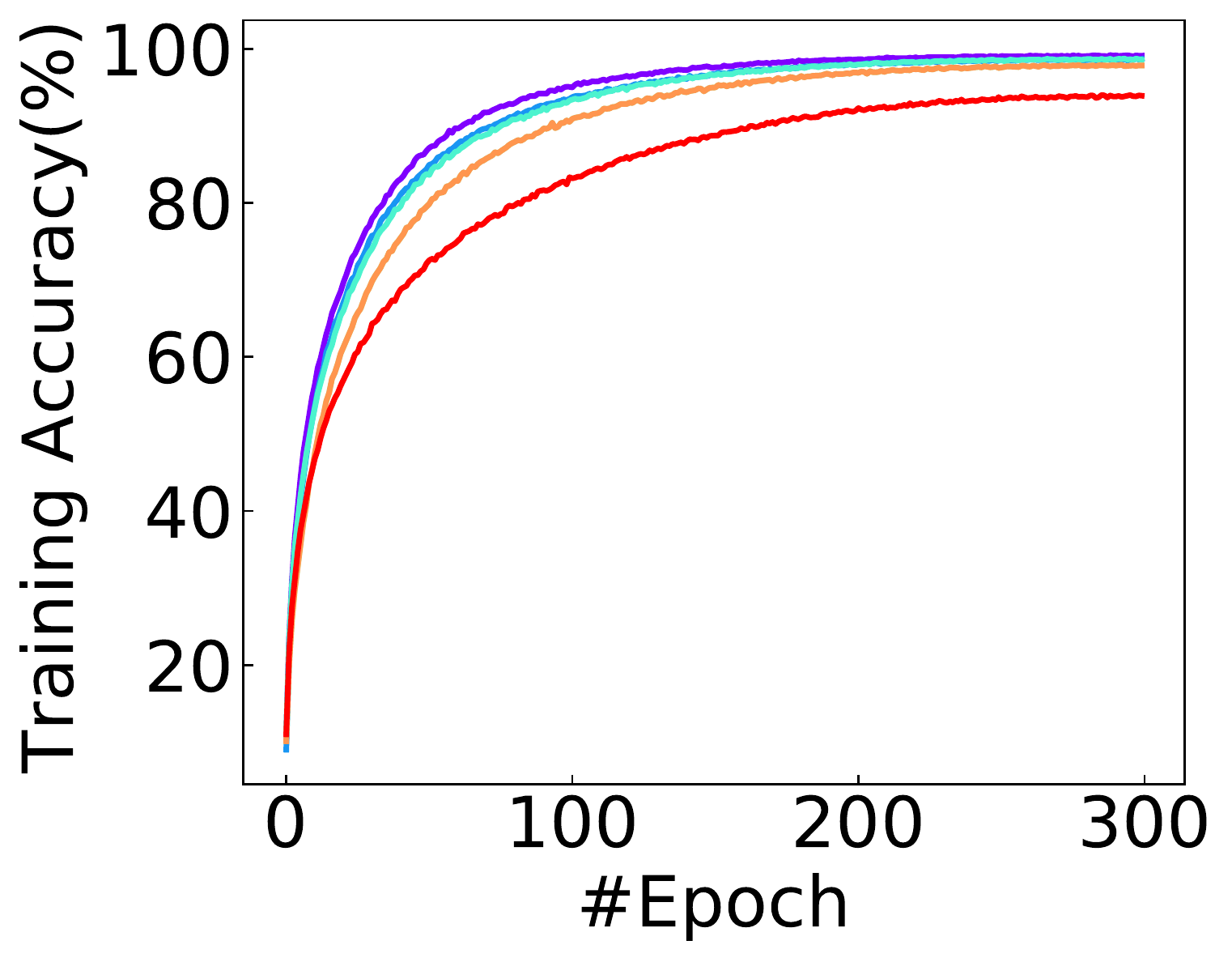} 
\\
{(a) } & { (b)} \\
\end{tabular}
\caption{ 
Comparison of  (a) training loss and (b) training accuracy across different searched architectures on CIFAR100.
}
\label{fig:method:training_convergence}
\end{center}
\end{figure}
%
\end{minipage}
%
%
\hfill
\begin{minipage}{0.48\textwidth}
\centering
\begin{figure}[H]
\begin{center}
\def\arraystretch{0.5}
\begin{tabular}{@{\hskip 0.00\linewidth}c@{\hskip 0.03\linewidth}c@{}c}
\includegraphics[width=0.475\linewidth]{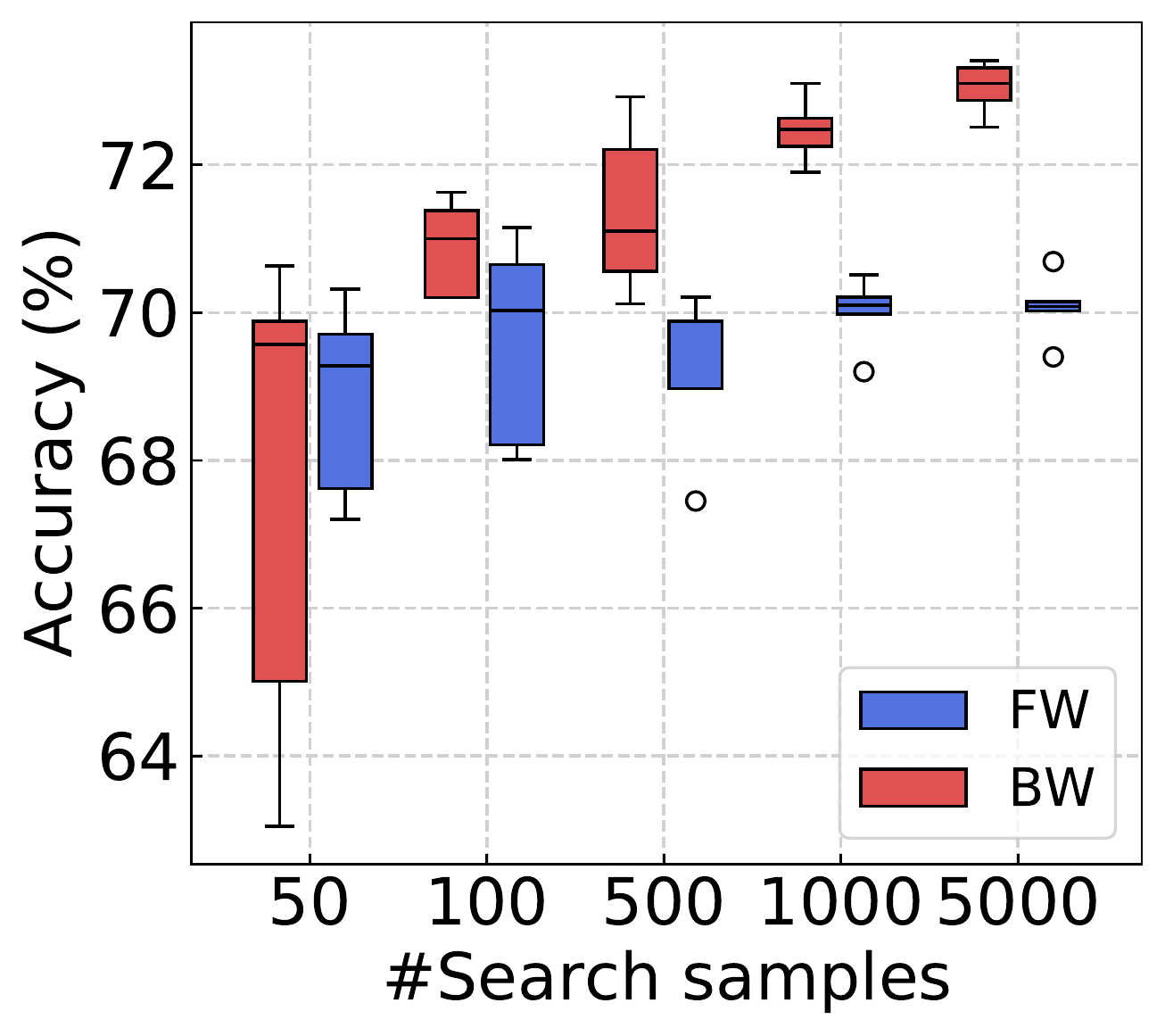} &
\includegraphics[width=0.46\linewidth]{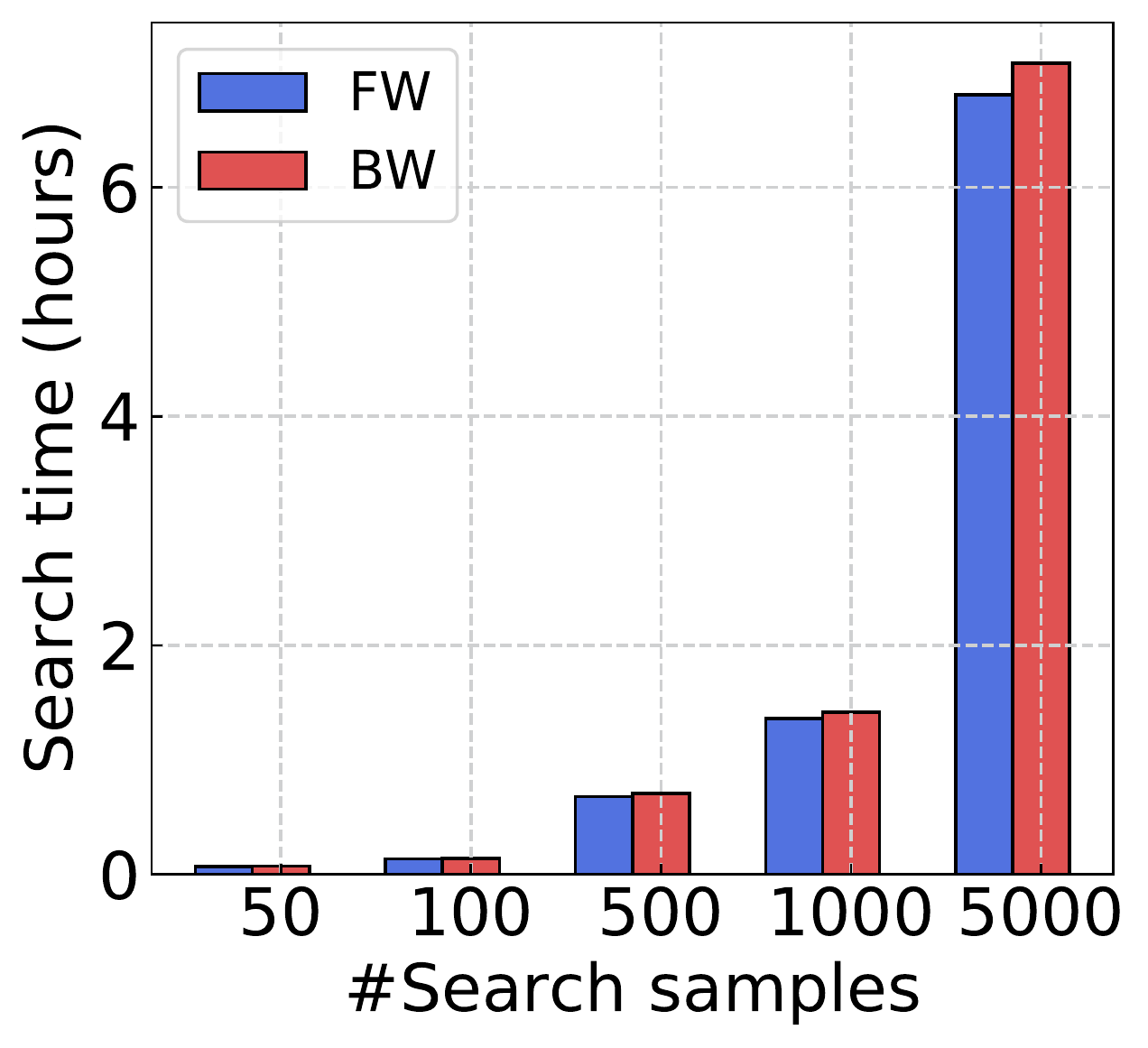} 
\\
{ (a) } & {(b)} \\
\end{tabular}
\caption{ (a) Accuracy and (b) search time with respect to number of search samples. We run the same settings 5 times. 
}
\label{fig:method:sample_and_timestep}
\end{center}
\end{figure}
\end{minipage}
\end{minipage}
%
\end{figure}

\noindent\textbf{Transferability of Searched Architecture.}
We conduct transferability analysis on searched SNN architecture in order to check the dependency of our searching method on the dataset.
We search the optimal architecture on dataset $A$ and train/test the searched architecture on dataset $B$ ($A \neq B$).
In Table \ref{table:exp:transferability}, for both forward and backward configurations, we use the searched architecture (Fig. \ref{fig:exp:best_architecturecell}) for all experiments. 
Note, $\Delta$Acc in \ref{table:exp:transferability} denotes the performance difference between the accuracy of the original searched architecture and transferred architecture.
The results show that the searched SNASNets are surprisingly transferable across diverse datasets, which opens up the promising advantage of eliminating searching time for huge and complex datasets.

\noindent\textbf{Analysis on Distance Metric.}
In our method, we use we propose Sparsity-Aware Hamming Distance (SAHD) where Hamming Distance (HD) is normalized based on the sparsity of the given two binary activation patterns.
This effectively addresses a large variation of spike activation across data samples.
We evaluate the performance of SNASNet according to the distance metric: HD vs. SAHD.
In Table \ref{table:exp:metric_ablation}, we report the performance of SNASNet-Fw and SNASNet-Bw on CIFAR100.
Here, we also run the search algorithm 5 times and report the mean and standard deviation.
The results demonstrate that SAHD reveals better architecture with less standard deviation in terms of test accuracy for both SNASNet-Fw and SNASNet-Bw architectures.

\noindent\textbf{Ablation on Number of Search Samples.}
In Fig. \ref{fig:method:sample_and_timestep}(a), we report the accuracy with respect to the number of search samples used in our searching algorithm.
We change the number of search samples [50, 100, 500, 1000, 5000] on CIFAR100.
The backward connection configuration (marked as red) shows higher variation as well as higher performance increase compared to that of the forward connection setting.
This is because searching backward connections has larger search space than searching forward connections only.
We also measure the computational time for searching in  Fig. \ref{fig:method:sample_and_timestep}(b).
We conduct the experiments on NVIDIA RTX 2080ti GPU and Intel(R) Xeon(R) Gold 6240 CPU @ 2.60GHz processor. 
The results show that searching backward connection requires sightly longer time than searching forward connection.

\noindent\textbf{Analysis on Timesteps.}
In Table \ref{table:exp:timestep_ablation}, we report the performance on CIFAR100 with respect to the number of timesteps used in SNNs.
Both SNASNet-Fw and SNASNet-Bw achieve performance gain with more number of timesteps.
SNASNet-Fw and SNASNet-Bw with 20 timesteps have improved accuracy by $0.48\%$ and $1.2\%$ compared to 5 timesteps, respectively.
Interestingly, the performance gain from SNASNet-Bw is larger than SNASNet-Fw.
The results suggest that adding backward connections to SNNs effectively leverages the temporal information for improved learning, and thus supports the advantage of backward connections in SNNs.

\begin{table}[t]
\noindent\begin{minipage}{\textwidth}
\begin{minipage}{0.48\textwidth}
\centering
\caption{Performance comparison between HD and  SAHD on CIFAR100. }

\resizebox{0.9\textwidth}{!}{%
\begin{tabular}{lcc}
\toprule
Architecture &   HD \:\:& SAHD  \\
\midrule
SNASNet-Fw \:\: &   64.16 $\pm$ 2.02\:\: &  70.06 $\pm$ 0.45 \\
  SNASNet-Bw\:\: &  66.80 $\pm$ 1.73 \:\:&  73.04 $\pm$ 0.36 \\
\bottomrule
\end{tabular}%
}
\label{table:exp:metric_ablation}
\end{minipage}
\hfill
\begin{minipage}{0.48\textwidth}
\centering
\caption{Ablation studies on timesteps on CIFAR100.}
 \resizebox{0.95\textwidth}{!}{%
\begin{tabular}{lcccc}
\toprule
Architecture\:\: &   T=5 \:\:& T=10\:\: & T=15\:\:& T=20\:\: \\
\midrule
SNASNet-Fw \:\: & 70.06 \:\:& 70.08\:\:& 70.56\:\:& 70.52\:\: \\
SNASNet-Bw \:\:& 73.04\:\:& 73.46\:\: & 73.49 \:\:& 74.24  \:\: \\
\bottomrule
\end{tabular}%
}
\label{table:exp:timestep_ablation}
\end{minipage}
\end{minipage}
\end{table}


\section{Conclusion}
In this paper, we search better SNN architecture using the temporal activation pattern of initialized network.
Our search space considers backward search connections in addition to forward connections, which brings the benefit of using temporal information.
By achieving better performance than the previous works, we demonstrate that a new type of architecture is more suitable for SNNs where spikes convey information through multiple timesteps. 

\noindent\textbf{Acknowledgment.}
This work was supported in part by C-BRIC, a JUMP center sponsored by DARPA and SRC, Google Research Scholar Award, the National Science Foundation (Grant\#1947826), TII (Abu Dhabi) and the DARPA AI Exploration (AIE) program.

\clearpage
%
%
\bibliographystyle{splncs04}
\bibliography{egbib,egbib_nas,egbib_actpattern}

\end{document}